\documentclass[aps,superscriptaddress,twocolumn,longbibliography]{revtex4-1}
\pdfoutput=1
\usepackage{amsmath}
\usepackage{amssymb}
\usepackage{graphicx}
\usepackage{xcolor}
\usepackage{braket}
\usepackage{verbatim}
\usepackage{marginnote}
\usepackage{mathtools}

\usepackage{hyperref}

\newcommand{\beq}{\begin{equation}}
\newcommand{\eeq}{\end{equation}}
\newcommand{\beqa}{\begin{eqnarray}}
\newcommand{\eeqa}{\end{eqnarray}}
\newcommand{\ben}{\begin{enumerate}}
\newcommand{\een}{\end{enumerate}}

\begin{document}

\title{Learning Relevant Features of Data with Multi-scale Tensor Networks}

\author{E.\ Miles Stoudenmire}
\affiliation{Center for Computational Quantum Physics, Flatiron Institute, 162 5th Avenue, New York, NY  10010, USA}

\date{\today}

\begin{abstract}
Inspired by coarse-graining approaches used in physics, we show how similar algorithms
can be adapted for data. The resulting algorithms are based on layered tree tensor
networks and scale linearly with both the dimension of the input and the training set size.
Computing most of the layers with an unsupervised algorithm, then optimizing just the top layer
for supervised classification of the MNIST and fashion-MNIST data sets gives very good results.
We also discuss mixing a prior guess for supervised weights together with an 
unsupervised representation of the data, yielding a smaller number of features 
nevertheless able to give good performance.
\end{abstract}

\maketitle
 

\section{Introduction}

Tensor decompositions are proving to be a powerful tool for machine learning, 
with applications in a wide variety of contexts from optimization 
algorithms~\cite{Anandkumar:2014,Sedghi:2016},
to compression of weight parameters~\cite{Novikov:2015,Cohen:2016,Novikov:2016,Stoudenmire:2016s,Yu:2017,Kossaifi:2017,Hallam:2017}, 
to theories of model expressivity and inductive bias~\cite{Cohen:2016,Cohen:2016i}. 
Tensors can parameterize complex combinations of more basic features \cite{Cohen:2016,Cohen:2016i,Blondel:2016,Novikov:2016,Stoudenmire:2016s}, and 
tensor decompositions can make these parameterizations efficient.
 
\begin{figure}[t]
\includegraphics[width=\columnwidth]{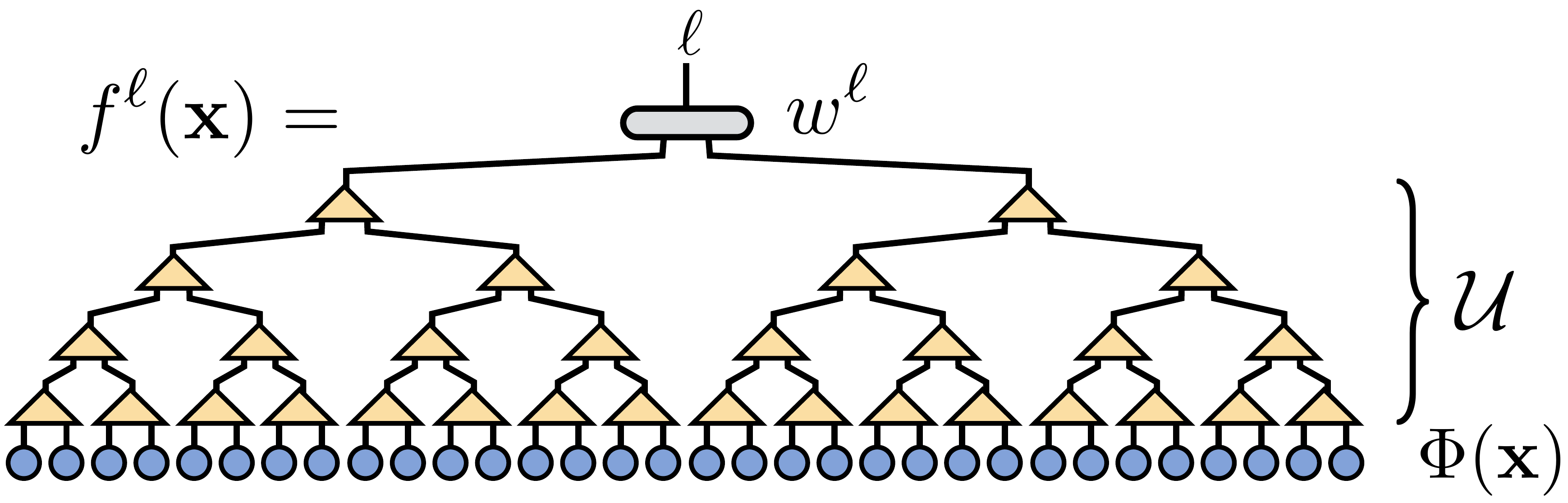}
\caption{The models $f^\ell(\mathbf{x})$ we will construct  
are defined by a tree tensor network that coarse grains the 
feature vector $\Phi(\mathbf{x})$ via the isometric tree tensor
layer $\mathcal{U}$. In Section~\ref{sec:unsupervised}, the layers $\mathcal{U}$
are determined using a purely unsupervised algorithm generalizable to various tasks. 
In Section~\ref{sec:mixed} the tensors in $\mathcal{U}$ are instead computed by a 
mixture of the unsupervised approach and an initial guess of a lower-accuracy solution for 
a supervised task.
After the coarse graining, the resulting tensors are contracted
with the top tensor $w^\ell$ to compute $f^\ell(\mathbf{x})$ 
(in Sec.~\ref{sec:partial} the top tensor is generalized to be a second type of tensor network). 
Only the top tensor is optimized for the specific task, such as supervised learning. In the multi-task
setting, the top tensor has an extra uncontracted index $\ell$ running over labels as shown
in the figure.}
\label{fig:treemodel}
\end{figure}

\begin{figure}[t]
\includegraphics[width=\columnwidth]{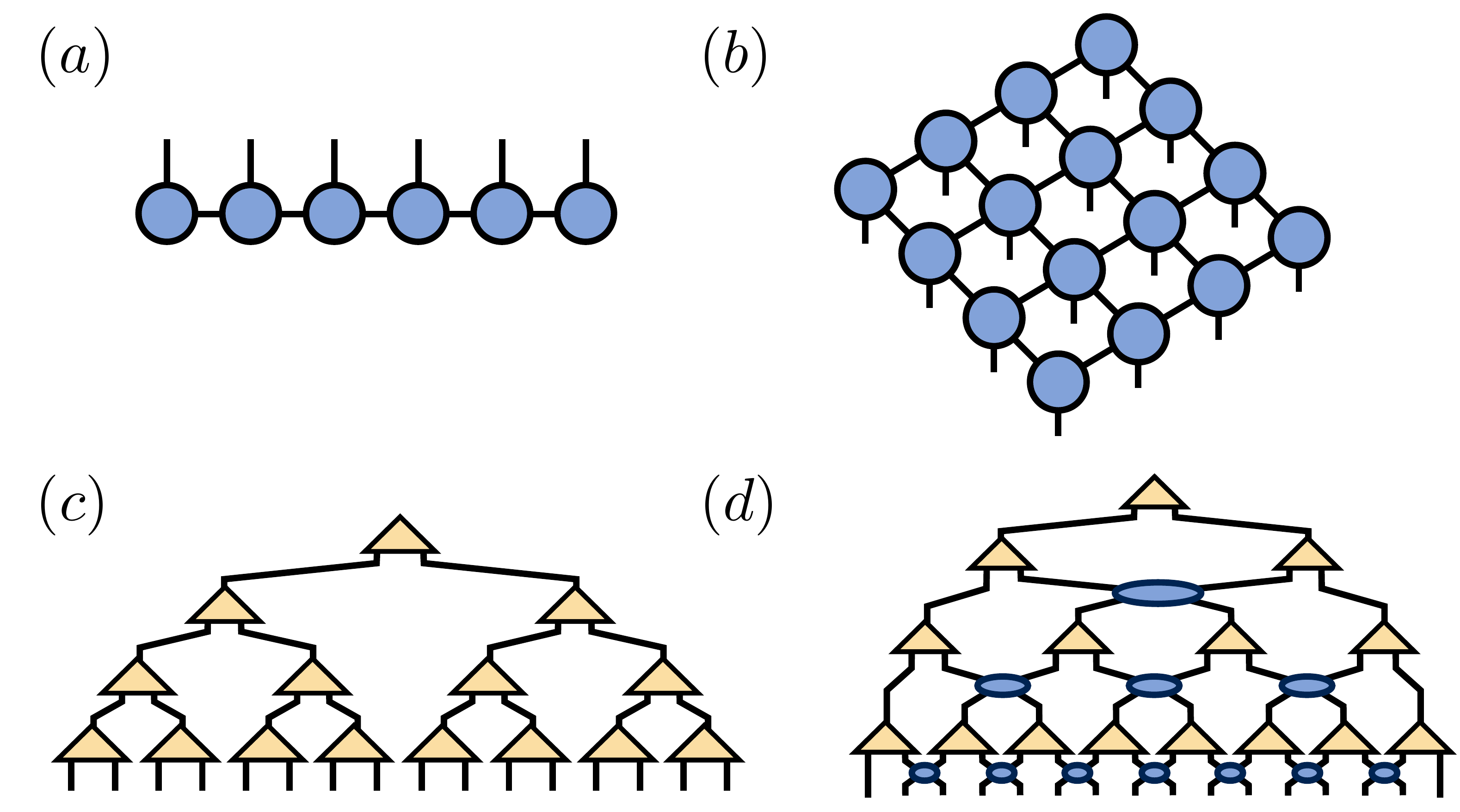}
\caption{Well studied tensor networks for compressing high-order tensors include 
the (a) matrix product state or tensor train network; (b) PEPS tensor network; 
(c) tree tensor or hierarchical Tucker network; and (d) MERA tensor network,
which is like a tree network but augmented with unitary \emph{disentangler} operations between branches
at each scale.}
\label{fig:tns}
\end{figure}

A particularly interesting class of tensor decompositions are \emph{tensor networks}.
These are factorizations of a very high-order tensor into a contracted network of 
low-order tensors---see Figs.~\ref{fig:treemodel},\ref{fig:tns}. Tensor networks break the curse of dimensionality by allowing 
operations such as contracting very high-order tensors or retrieving their components 
to be accomplished with polynomial cost by manipulating the low-order factor tensors \cite{Orus:2014a}.
 
A key motivation for the development of tensor networks was the idea of coarse graining. Known in physics as the \emph{renormalization group}, coarse graining is a way to gain insight into a complicated statistical system by marginalizing over its smallest length scales while preserving properties at larger scales \cite{Wilson:1971,Wilson:1979}. 
After each round of coarse graining, 
one identifies a new smallest scale and repeats the process in a hierarchical fashion. A related idea is a wavelet multiresolution analysis in applied mathematics \cite{Evenbly:2016w}. For the case of deep neural networks, there is evidence that certain networks implement coarse graining, with each layer of neurons learning progressively coarser features \cite{Erhan:2009}. 
Such an analogy between the renormalization group and neural nets has been made precise for the case of deep belief networks \cite{Mehta:2014}. Renormalization group ideas could yield insights into
 data when applied in the framework of PCA \cite{Bradde:2016}.

Numerical implementations of the renormalization group motivated the development of 
tensor networks \cite{Fannes:1992,Ostlund:1995,Vidal:2003,Verstraete:2004p,Vidal:2007} which not only underpin powerful algorithms for solving quantum 
\cite{White:1992,Schollwoeck:2005,McCulloch:2007,McCulloch:2008,Schollwoeck:2011} and classical \cite{Levin:2007,Evenbly:2014} 
systems in physics, but lead to the insight that the solution has a structure imparted
by the coarse-graining procedure itself. This structure is most striking in the case of the 
MERA family of tensor 
networks Fig.~\ref{fig:tns}(d) \cite{Vidal:2007,Evenbly:2009}, where the quantum wavefunction acquires an emergent 
extra dimension, or layered structure, whose geometry is reflective of the correlations of the system 
\cite{Evenbly:2011g,Swingle:2012m,Nozaki:2012,Hayden:2016}.

Recently a framework has been discussed by several groups \cite{Cohen:2016,Novikov:2016,Stoudenmire:2016s} 
that offers a particularly straightforward way to apply tensor network algorithms 
to machine learning tasks, such as supervised 
\cite{Novikov:2016,Stoudenmire:2016s,Levine:2017,Liu:2017,Khrulkov:2017}
and unsupervised \cite{Han:2017} learning. This framework can be viewed equivalently as a 
neural network architecture with linear activation and product pooling
 \cite{Cohen:2016,Cohen:2016c,Cohen:2016i,Levine:2017} or an approach to kernel learning
with the weights represented as a tensor network \cite{Novikov:2016,Stoudenmire:2016s}.
We will take the kernel learning perspective here, but nevertheless construct models
resembling deep neural networks.

When parameterizing a kernel learning model with a tensor network, 
all operations other than the initial feature map
are \emph{linear} when viewed as transformations on the entire feature space, 
which is exponentially higher dimensional than the input space. The 
fact that tensor networks are composed from linear maps is what makes them 
amenable to theoretical analysis \cite{Perez-Garcia:2007,Schuch:2010p,Evenbly:2011g} 
and useful for devising algorithms \cite{McCulloch:2007,Orus:2008,Chan:2008a}. In the context of parameterizing distributions, many 
interesting tensor networks are tractable and can be directly sampled \cite{Ferris:2012,Han:2017}.
Sums and products of tensor networks can be controllably approximated as a tensor 
network of the same type \cite{McCulloch:2007,Stoudenmire:2010}.
Tensor network representations of distributions can be proven to have exponentially 
or power-law decaying correlations depending on their geometry \cite{Evenbly:2011}.
Algorithms to optimize tensor networks are often adaptive, allowing dimensions of
internal indices to adjust as needed \cite{Schollwoeck:2011}.

In what follows we will use tensor network coarse graining 
to compress data originally represented in a very high dimensional space. 
The approach is unsupervised, based purely on statistical properties of the data.
Each step has a controlled accuracy, and the method is scalable to large 
data set sizes and input dimensions.
The resulting reduced description takes the form of a layered tensor network with a tree 
structure---a tree tensor network Fig.~\ref{fig:tns}(c)---and can be used to obtain good results on
 learning tasks by optimizing tensors only at the top layer.
One could further specialize all of the tensors in the network for a specific task,
but we will not do so here.

The cost of training each of the models discussed below 
is linear in both training set size and input dimension, 
assuming a fixed number of parameters.
The cost of evaluating the model on a test input is independent of 
training set size. The scaling of the optimization algorithm with training set size could 
be reduced to sub-linear with stochastic optimization techniques.

Throughout we will use tensor diagram notation. For a brief introduction to this notation,
see the Appendix. The experiments were implemented using the ITensor 
software \cite{ITensor}.

\section{Motivation and Background}

The algorithm we will develop is motivated by the fact that within kernel learning, 
the optimal weights belong to the span of the training data within feature space.
More specifically, consider a model
\begin{align}
f(\mathbf{x}) = W\cdot\Phi(\mathbf{x})
\end{align}
defined in terms of a high-dimensional feature map $\Phi$ and weights $W$.
Given a set of training inputs $\{\mathbf{x}_j\}_{j=1}^{N_T}$, it can be shown 
that for a broad set of learning tasks the optimal weights 
have the form
\begin{align}
W = \sum_{j=1}^{N_T} \alpha_j \Phi^\dagger(\mathbf{x}_j) \label{eqn:representer}
\end{align}
where only the $\alpha_j$ parameters remain to be optimized for the specific task.
The well-known fact that $W$ can be expressed this way is called the 
\emph{representer theorem}, which applies to many
common supervised tasks, as well as certain unsupervised tasks such as kernel 
PCA \cite{Scholkopf:2001}.

\subsection{Alternate Basis for Weight Parameters}

Parameterizing the weights $W$ by a set of numbers $\alpha_j$
of the size of the training set is a major improvement compared to
representing $W$ within the possibly infinite dimensional feature space defined by 
$\Phi$. But standard approaches to optimizing $W$ in terms of the $\alpha_j$ 
 typically exhibit quadratic or worse dependence on training set size, 
which can be prohibitive for state-of-the-art tasks with training sets
reaching millions in size.

To make progress, notice that the main content of Eq.~(\ref{eqn:representer}) is that
$W$ resides in the span of the $\{\Phi^\dagger(\mathbf{x}_j)\}$. Thus the optimal
weights $W$ can be expanded in any basis of vectors $U^\dagger_n$
\begin{align}
W = \sum_n \beta_n U^\dagger_n
\end{align}
as long as the $U^\dagger_n$ span the same space as the $\Phi^\dagger(\mathbf{x}_j)$.

If cost was no concern, a natural way to obtain a basis $U$ would be to
think of the  training set feature vectors as a matrix
$\Phi_j^{\mathbf{s}}=\Phi^{s}(\mathbf{x}_j)$ and obtain $U$ from 
singular value decomposition of $\Phi_j^{\mathbf{s}}$ as
\begin{align}
\Phi_j^{\mathbf{s}} = \sum_{n n'} U^{\mathbf{s}}_n S^{n}_{n'} V^{\dagger n'}_j \ .
\end{align}
where $S$ is the diagonal matrix of singular values \mbox{$s_n = S^{n}_{n}$}.
By inserting this decomposition into Eq.~(\ref{eqn:representer}), 
the optimal weights $W$ can indeed be written
\begin{align}
W_\mathbf{s} = \sum_{j n n'} \alpha_j V^j_{n'} S^{n'}_{n} \,U^{\dagger n}_ \mathbf{s}
= \sum_n \beta_n \,U^{\dagger n}_\mathbf{s}  \label{eqn:Uform}
\end{align}
which explicitly relates the $\alpha_j$ and $\beta_n$ parameters.

One advantage of expressing $W$ in terms of $U$ is that the columns of $U$ are
orthonormal. But more importantly, if the dimension of feature space is much larger 
than the minimum statistically significant training set size,  
then many singular values $s_n$ will be very small or zero and the corresponding
rows of $U^\dagger$ can be discarded. Following such a truncation, 
Eq.~(\ref{eqn:Uform}) says that to a good approximation, the optimal weights
can be parameterized within a significantly reduced space of parameters $\beta_n$ and 
$U^\dagger$ is the transformation from the entire feature space to the reduced
parameter space.

Computing the singular value decomposition of $\Phi_j^{\mathbf{s}}$ directly would not
scale well for large training sets or high-dimensional feature maps, yet as we will show 
it is nevertheless possible to efficiently determine the
transformation $U$ in truncated form. Observe that $U$ diagonalizes the 
\emph{feature space covariance matrix} \cite{Scholkopf:1998} defined as 
\begin{align}
\rho^{\mathbf{s}'}_\mathbf{s} & = \frac{1}{N_T} \sum_{j=1}^{N_T} \Phi^{\mathbf{s}'}_j \Phi^{\dagger j}_\mathbf{s} \\
& = \sum_{n} U^{\mathbf{s}'}_{n}\,  P_n \, U^{\dagger n}_{\mathbf{s}} 
\end{align}
where $P_n = (S^{n}_{n})^2$ are the eigenvalues of the Hermitian matrix $\rho$.
As we demonstrate in Sec.~\ref{sec:unsupervised} below, 
the feature space covariance matrix $\rho$ is amenable to 
decomposition as a layered \emph{tensor network}. Computing every layer of this network
can provide an efficient expression for the elements of the basis $U$ corresponding
to the largest eigenvalues of $\rho$. Computing only some of the layers still 
has the beneficial effect of projecting out directions in feature space along which $\rho$
has small or zero eigenvalues. By carrying out an iterative procedure 
to truncate directions in feature space along which $\rho$ has a very small projection, 
one can rapidly reduce the size of the space needed to carry out learning tasks.

\begin{figure}[t]
\includegraphics[width=0.85\columnwidth]{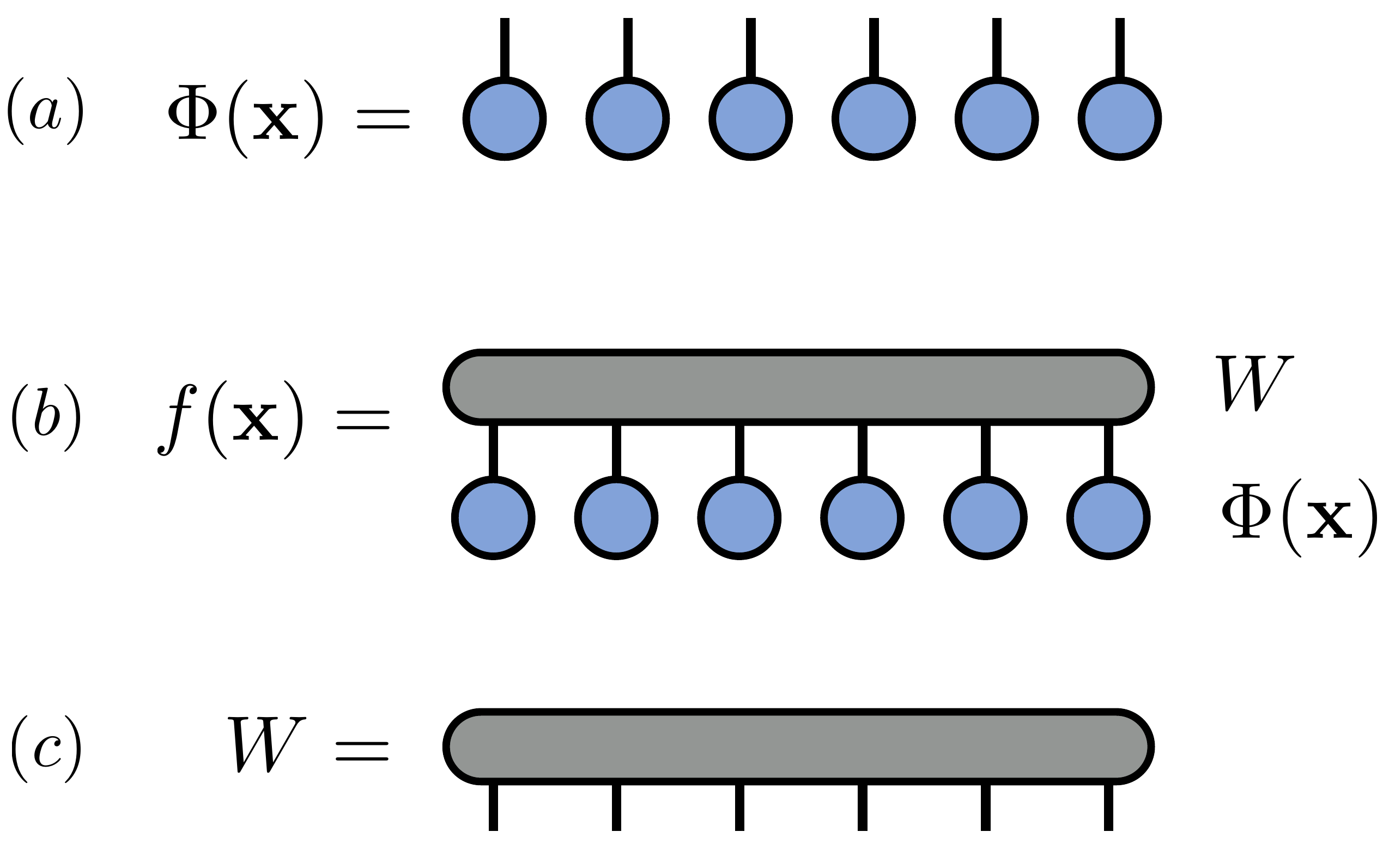}
\caption{Choosing the feature map (a) to be a tensor product of local feature maps
leads to a model $f(\mathbf{x})$ of the form (b) where the weight parameters $W$ have (c) the structure
of an order-N tensor.}
\label{fig:feature_map}
\end{figure}

We will also see that $\rho$ is not the only choice of matrix for determining
a tensor network basis for features.
As demonstrated in Sec.~\ref{sec:mixed},
other choices result in a network more adapted for a specific task, and can 
have fewer latent parameters without reducing model performance.

\subsection{Tensor Product Feature Maps}

Before describing the algorithm to partially or fully diagonalize $\rho$ as a tensor network,
we briefly review the class of feature maps 
which lead to a natural representation of model parameters as a tensor network,
as discussed in Refs.~\onlinecite{Cohen:2016,Novikov:2016,Stoudenmire:2016s}.
These are feature maps $\Phi(\mathbf{x})$ which map inputs $\mathbf{x}$ from a space of dimension $N$ into a space of dimension $d^N$ with a tensor product structure. The simplest case of 
such a map begins by defining a \emph{local feature map} $\phi^{s_j}(x_j)$ where $s_j=1,2,\ldots,d$. 
These local feature maps define the full feature map as:
\begin{align}
\Phi^{s_1 s_2 \cdots s_N}(\mathbf{x}) = \phi^{s_1}(x_1) \phi^{s_2}(x_2) \cdots \phi^{s_N}(x_N)
\label{eqn:Phi}
\end{align}
as shown in Fig.~\ref{fig:feature_map}(a), where placement of tensors next to each other implies
an outer product.
This choice of feature map leads to models of the form
\begin{align}
f(\mathbf{x}) = \sum_{s_1 s_2 \cdots s_N} W_{s_1 s_2 \cdots s_N} \phi^{s_1}(x_1) \phi^{s_2}(x_2) \cdots \phi^{s_N}(x_N) \label{eqn:full_model}
\end{align}
which are depicted in Fig.~\ref{fig:feature_map}(b).
As evident from the above expression, the weight parameters are indexed by $N$ indices of dimension $d$.
Thus there are $d^N$ weight parameters and $W$ is a tensor of order $N$. We will be interested in the 
case where $d$ is small (of order one or ten) and $N$ is many hundreds or thousands in size.

Of course, manipulating or even storing $d^N$ parameters quickly becomes impossible as $N$ increases.
A solution that is both practical and interesting is to assume that the optimal weights $W$
can be efficiently approximated by a \emph{tensor network} \cite{Orus:2014a,Schollwoeck:2011},
an idea proposed recently by several groups \cite{Cohen:2016, Novikov:2016,Stoudenmire:2016s,Levine:2017}.

A tensor network is a factorization of an order $N$
tensor into the contracted product of low-order tensors. 
Key examples of well-understood tensor networks for which efficient algorithms are known 
are depicted in Fig.~\ref{fig:tns} and include:
\begin{itemize} 
\item the matrix product 
state (MPS) \cite{Fannes:1992,Ostlund:1995,Vidal:2003,Banuls:2008} or tensor train decomposition \cite{Oseledets:2011}, Fig.~\ref{fig:tns}(a)
\item the PEPS tensor network \cite{Verstraete:2004p}, Fig.~\ref{fig:tns}(b)
\item the tree tensor network \cite{Friedman:1997,Shi:2006} or hierarchical Tucker decomposition \cite{Hackbusch:2009}, Fig.~\ref{fig:tns}(c)
\item the MERA tensor network \cite{Vidal:2007, Evenbly:2009}, Fig.~\ref{fig:tns}(d).
\end{itemize}
Each of these networks makes various tradeoffs in terms of how complicated they are to manipulate
versus their ability to represent statistical systems with
higher-dimensional interactions or more slowly decaying correlations.
A good introduction to tensor networks in the physics context is 
given by Or\'{u}s in Ref.~\onlinecite{Orus:2014a} and in a mathematics context by 
Cichocki in Ref.~\onlinecite{Cichocki:2014}. Other detailed reviews include 
Refs.~\onlinecite{Schollwoeck:2011,Evenbly:2011g,Bridgeman:2016,Silvi:2017}.

\begin{figure}[b]
\includegraphics[width=\columnwidth]{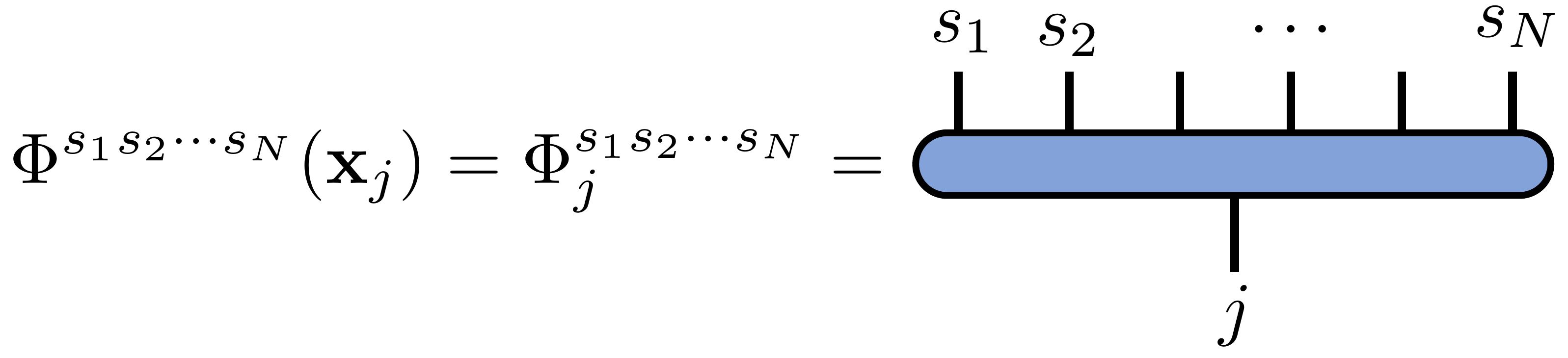}
\caption{It is sometimes convenient to view the collection of 
feature vectors $\Phi(\mathbf{x}_j)$ indexed over each training input
$\mathbf{x}_j$ as a single tensor.}
\label{fig:phi_tensor}
\end{figure}

\section{Unsupervised Coarse Graining \label{sec:unsupervised}}

As discussed in the previous section, if one can compute the eigenvectors $U^\mathbf{s}_n$
of the feature space covariance matrix $\rho$, defined as
\begin{align}
\rho & = \frac{1}{N_T} \sum_{j=1}^{N_T} \Phi(\mathbf{x}_j) \Phi(\mathbf{x}_j)^\dagger \label{eqn:rho} \\
& = \sum_{n} U^{\mathbf{s}'}_{n}\,  P_n \, U^{\dagger n}_{\mathbf{s}} 
\end{align}
then the optimal weights for a wide variety of learning tasks can be represented as
\begin{align}
W_\mathbf{s} = \sum_n \beta_n \,U^{\dagger n}_\mathbf{s} \ . 
\end{align}
Furthermore, for a specific task
the eigenvectors of $\rho$ with small enough eigenvalues can be discarded without reducing performance.
As an example, for the task of supervised learning with a quadratic cost, if the cost function includes a quadratic weight penalty $\lambda |W|^2$, eigenvectors whose eigenvalues are much smaller than $\lambda$
do not contribute significantly to $W$ and can be projected out.

\begin{figure}[t]
\includegraphics[width=0.85\columnwidth]{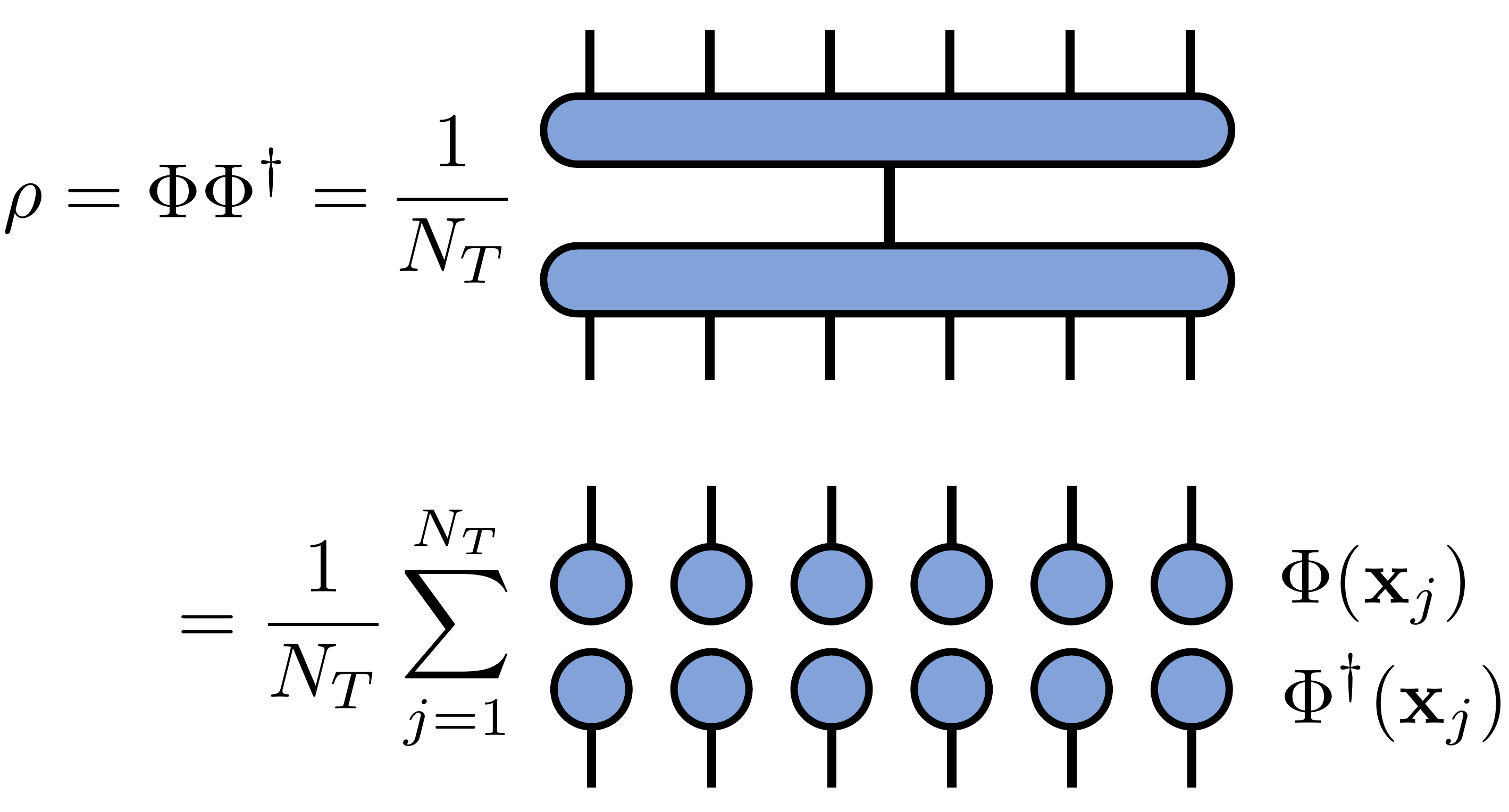}
\caption{The feature space covariance matrix can be formed by contracting two copies of $\Phi$
over the training set index~$j$. In practice one does not form this entire object but uses
efficient algorithms to compute reduced covariance matrices.}
\label{fig:rho}
\end{figure}

Now we will outline a strategy to find a controlled approximation for the dominant eigenvectors 
of $\rho$ using a layered tensor network. For the purposes of this section, 
it will be useful to think of 
\begin{align}
\Phi^{s_1 s_2 \cdots s_N}(\mathbf{x}_j) = \Phi^{s_1 s_2 \cdots s_N}_j = \Phi^{\mathbf{s}}_j
\end{align}
as a tensor of order $(N+1)$ as shown in Fig.~\ref{fig:phi_tensor}. In this view, $\rho$ is 
formed by contracting $\Phi$ and $\Phi^\dagger$ over the training data index $j$ as shown in Fig.~\ref{fig:rho}.

\begin{figure}[t]
\includegraphics[width=0.7\columnwidth]{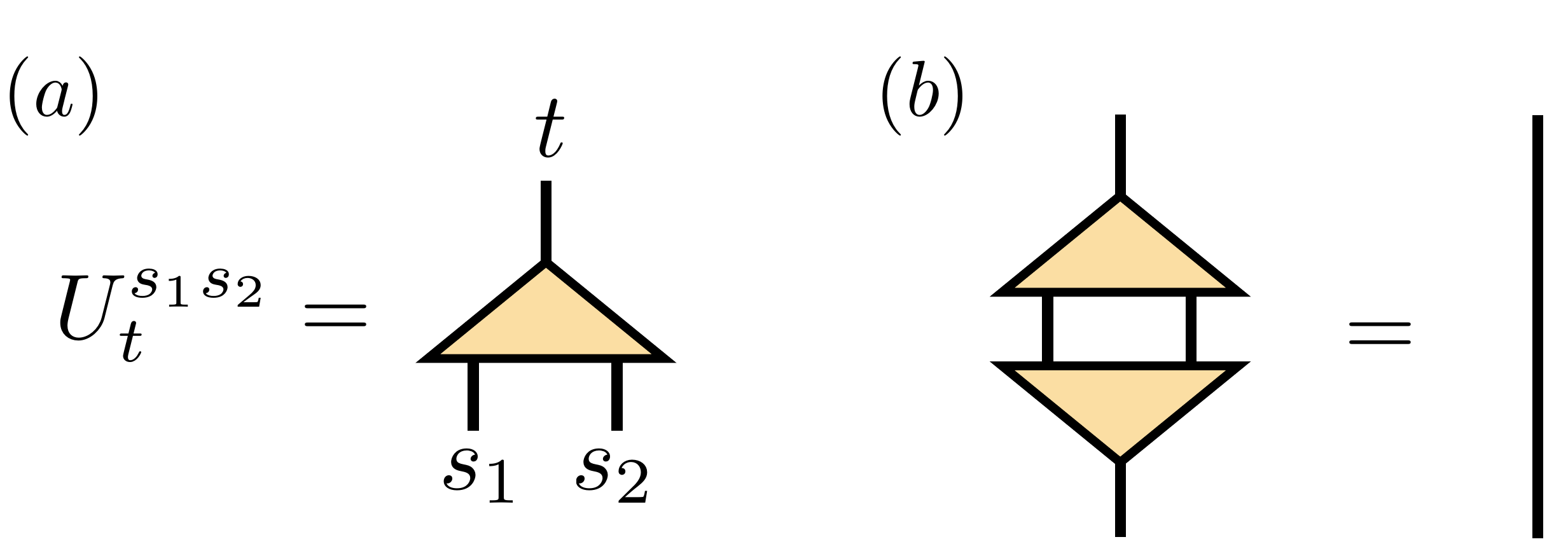}
\caption{An tensor (a) $U_t^{s_1 s_2}$ which is an isometry mapping two vector spaces whose bases are 
labeled by $s_1$ and $s_2$ into a single vector space labeled by $t$ obeys the 
condition that (b) contracting $U$ and $U^\dagger$ over the $s_1$ and $s_2$ indices
yields the identity matrix (represented diagrammatically as a line).}
\label{fig:isom}
\end{figure}

\begin{figure}[b]
\includegraphics[width=0.6\columnwidth]{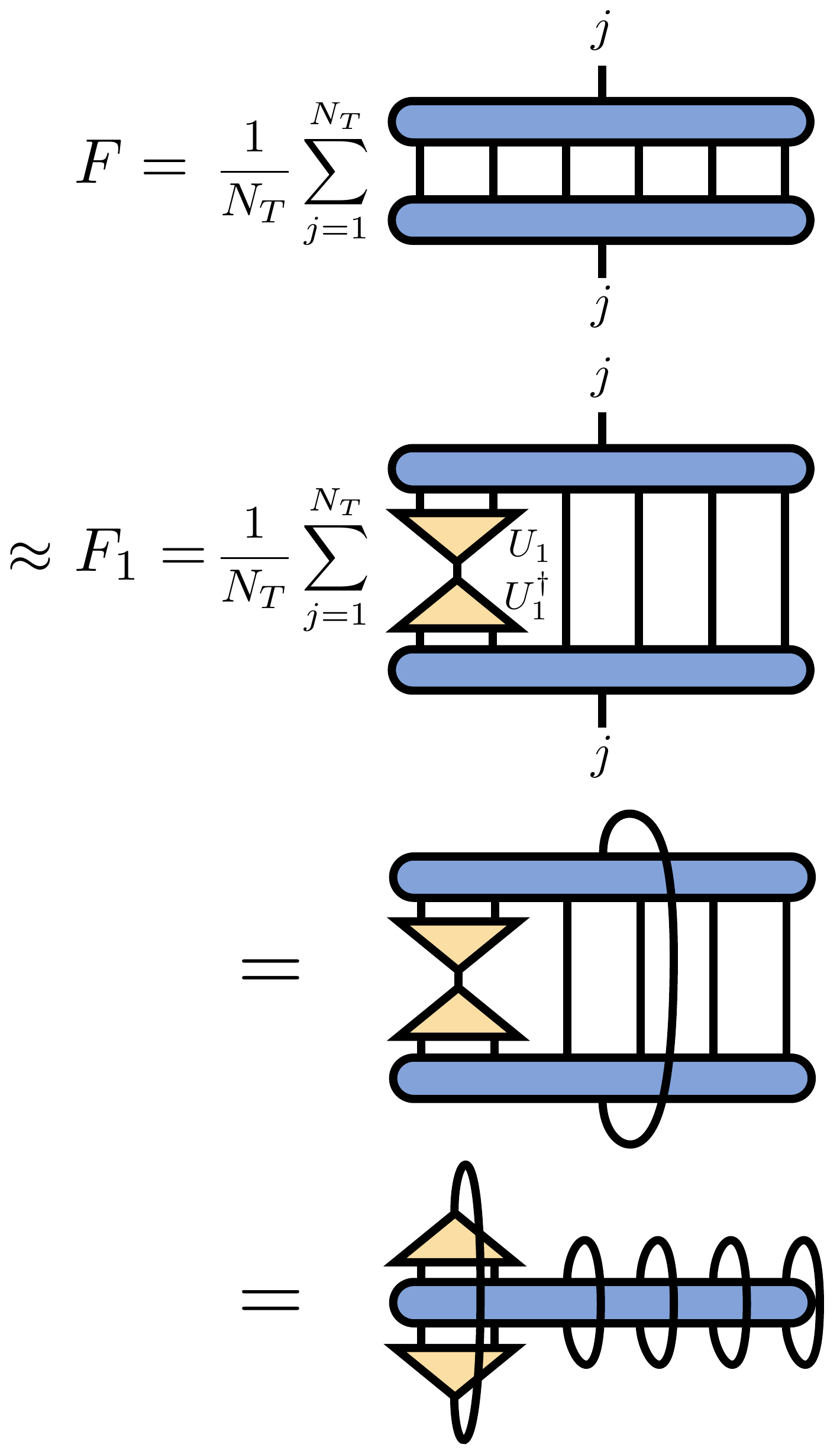}
\caption{The fidelity $F$ is defined as the average inner product of the training 
set feature vectors, or equivalently the trace of the covariance matrix $\rho$.
The isometry $U_1$ is chosen to maximize the fidelity following coarse graining
(second panel above) which is equivalent to maximizing the trace of 
 $\rho_{12}$ after coarse graining (last panel above).}
\label{fig:f12}
\end{figure}

Because it is not feasible to diagonalize $\rho$ directly, the strategy we will pursue is 
to compute local \emph{isometries} which combine two indices
into one and project out subspaces of the feature space spanned by eigenvectors of $\rho$
with small eigenvalues, as defined by some pre-defined cutoff or threshold $\epsilon$.
The term \emph{isometry} here refers to a third-order tensor $U^{s_1 s_2}_t$ such that
\begin{align}
\sum_{s_1 s_2} U^{s_1 s_2}_t U_{s_1 s_2}^{t'} = \delta^{t'}_{t}
\end{align}
where $\delta^{t'}_{t}$ is the Kronecker delta tensor (or identity matrix).
This isometric constraint is depicted in Fig.~\ref{fig:isom}(b).
The dimension of $t$ can be less than or equal to the product of the dimensions
of $s_1$ and $s_2$. 
The isometric property of $U$ means it can be interpreted as a unitary rotation followed by a projection.
This straightforward interpretation could be useful for interpreting and analyzing a learned model
after training.

To compute the first isometry $U_1$ in the network we want to construct,
define $U_1$ such that when it acts on the first two feature space indices
$s_1$ and $s_2$ of the tensor $\Phi^{\mathbf{s}}_j$ it maximizes the fidelity 
$F$, defined as
\begin{align}
F = \text{Tr}[\rho] = \frac{1}{N_T} \sum_j \Phi^\dagger_j \Phi_j \ .
\end{align}
After coarse graining the feature map using the isometry $U_1$ 
the fidelity of the resulting approximation to $\rho$ is
\begin{align}
F_1 = \frac{1}{N_T} \sum_j \Phi^\dagger_j U_1 U_1^\dagger \Phi_j
\end{align}
as shown in Fig.~\ref{fig:f12}. 
Because $U_1$ is an isometry, the fidelity $F_1$ of the coarse-grained
feature vectors $U_1^\dagger \Phi_j$ is always less than or equal to $F$.
The fidelity will be maximized if $U_1$ projects out a subspace 
of the full feature space within which the components of the feature vectors
$\Phi^{\mathbf{s}}_j$ are on average very small.

\begin{figure}[t]
\includegraphics[width=0.85\columnwidth]{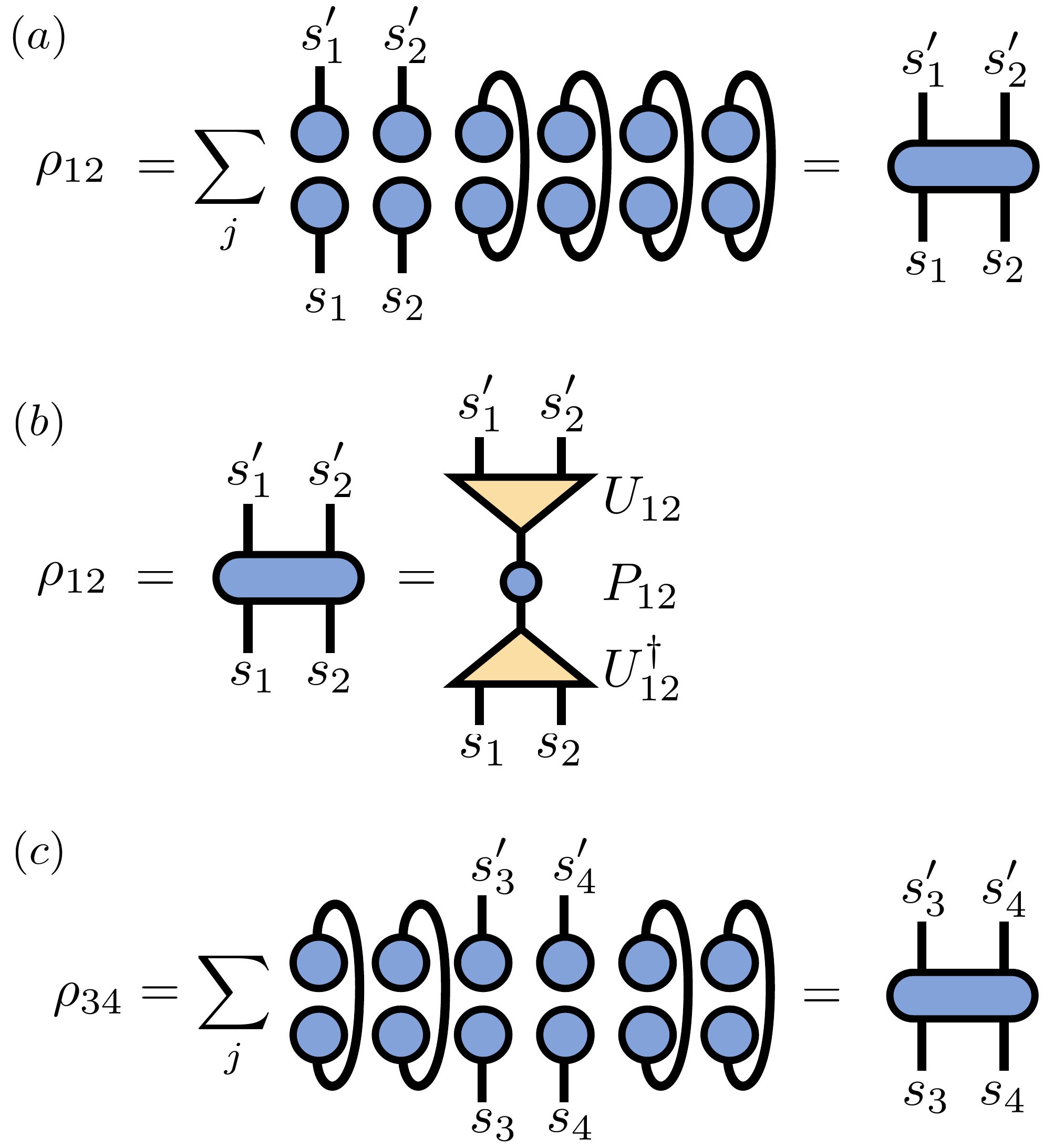}
\caption{Definition (a) of the reduced covariance matrix $\rho_{12}$; (b) computation
of the optimal isometry $U_1$ by diagonalizing $\rho_{12}$ and truncating its smallest
eigenvalues; (c) definition of the reduced covariance matrix $\rho_{34}$.}
\label{fig:rho12}
\end{figure}

To solve for the optimal isometry $U_1$, it is convenient to introduce
the \emph{reduced covariance matrix} $\rho_{12}$, defined by
tracing over all of the indices $s_3,s_4,\ldots,s_N$ of $\rho$
as shown in Fig.~\ref{fig:rho12}(a). The motivation for introducing the 
reduced covariance matrix can be seen in the last two panels of 
Fig.~\ref{fig:f12}. The manipulations there show the fidelity
$F_1$ can be written in terms of $\rho_{12}$ as
\begin{align}
F_1 = \sum_{s_1 s_2 s'_1 s'_2 t} U_1^{\dagger\, t}\!_{s'_1 s'_2}\, \rho_{12}^{s'_1 s'_2}\,_{s_1 s_2} U_1^{s_1 s_2}\!_{t} \ .
\end{align}
It follows that the optimal isometry $U_1$ can be computed by diagonalizing $\rho_{12}$ as 
\begin{align} 
\rho_{12} = U_1 P_{12} U_1^\dagger \, ,
\end{align}
here viewing $\rho_{12}$ as a matrix with row index $(s_1 s_2)$ and column index $(s'_1 s'_2)$ as shown in Fig.~\ref{fig:rho12}. The matrix $P_{12}$ is a diagonal matrix whose diagonal 
elements are the eigenvalues of $\rho_{12}$.
After the diagonalization, 
the columns of $U_1$ are chosen to be the eigenvectors corresponding to the $D$ largest eigenvalues
of $\rho_{12}$. Let the rank of the matrix $\rho_{12}$ be $r$ and call its eigenvalues
$\{ p_i \}_{i=1}^{r}$. 
One way to determine the number $D$ of eigenvalues to keep is to 
choose a predetermined threshold $\epsilon$ and define $D$ such that the 
\emph{truncation error} $E$ is less than $\epsilon$, where the truncation error is defined as
\begin{align}
E = \frac{\sum_{i=D}^{r} p_i}{\text{Tr}[\rho_{12}]} < \epsilon \ . \label{eqn:truncerr}
\end{align}
This is the same procedure proposed by White in the context of the density matrix
renormalization group (DMRG) algorithm used in quantum mechanics,
where the $\Phi^{\mathbf{s}}_j$ is analogous to an ensemble of wavefunctions 
enumerated by $j$; $\rho$ is the full density matrix;
and $\rho_{12}$ is a reduced density matrix \cite{White:1992}.

\begin{figure}[b]
\includegraphics[width=0.8\columnwidth]{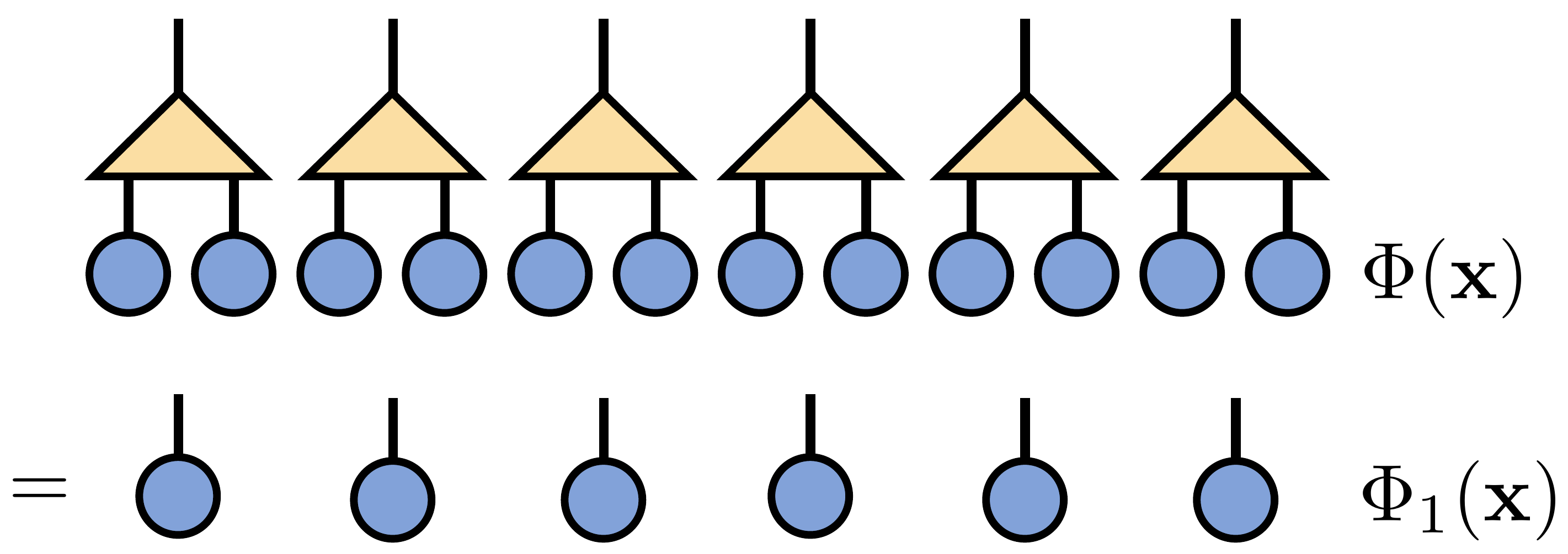}
\caption{Having determined a layer of isometries, these isometries can
be used to coarse grain the feature vectors.}
\label{fig:rg}
\end{figure}

To compute the remaining isometries which will form the first layer, 
the procedure continues in an analogous
fashion by next computing the reduced covariance matrix $\rho_{34}$ as in Fig.~\ref{fig:rho12}(c)
and diagonalizing it to obtain the isometry $U_{34}$. 
Note that the calculation of the reduced covariance matrices as well as the individual summations over the training data used to produce them can be performed in parallel. What is more,
we find that when summing over the training data in a random order, 
the reduced covariance matrices typically converge before summing over the entire training
set, and this convergence can be monitored to ensure a controlled approximation.
After diagonalizing the reduced covariance matrices for every pair of local indices $(s_{2i-1}, s_{2i})$, 
one obtains the first layer of isometries depicted in Fig.~\ref{fig:rg}.

The isometry layer can now be used to coarse grain each of the training set feature vectors
$\Phi(\mathbf{x}_j)$. After the coarse graining, one can repeat the process again to make a second layer
by making a new covariance matrix in the coarse grained feature space and diagonalizing its reduced
covariance matrices.

\begin{figure}[t]
\includegraphics[width=0.9\columnwidth]{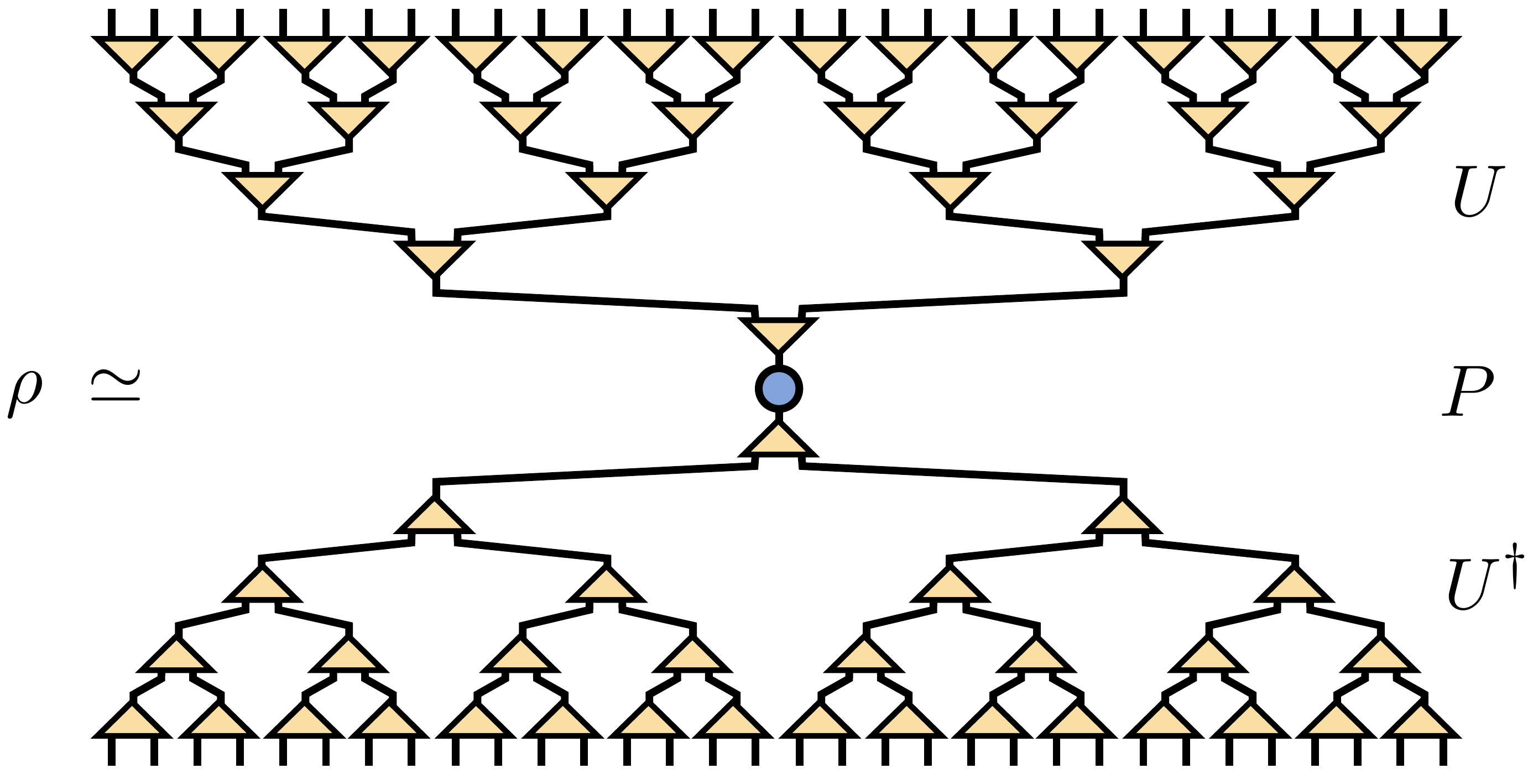}
\caption{Iterating the algorithm of Sec.~\ref{sec:unsupervised} for $\log_2(N)$ steps
approximately diagonalizes $\rho=UPU^\dagger$ with the diagonalizing unitary $U$ approximated
as a tree tensor network. The error made in the approximation is controlled
by the truncation errors made at each step.}
\label{fig:diagrho}
\end{figure}

If the coarse-graining procedure is repeated $\log_2(N)$ times 
(assuming $N$ is a power of two), the end result is that one has 
approximately diagonalized $\rho$ in the form shown in Fig.~\ref{fig:diagrho}, with $U$ expressed
in tree tensor network form.
With the particular choice of $\rho$ made above, the resulting algorithm is an approximate
implementation of kernel principal component analysis (kernel PCA) \cite{Scholkopf:2001}, but computed directly in feature space, not in the dual formulation. If one made no truncations 
throughout the algorithm, the equivalence to kernel PCA would be exact, but in practice
the dominant eigenvectors of $\rho$ are approximated by a tensor network with smaller
internal indices than needed to capture $U$ exactly. 
The fact that the calculation can be performed efficiently 
results from both the choice of a feature map with low-rank structure Eq.~\ref{eqn:Phi}
and on the iterative algorithm based on tree tensor networks.
But whether the approximation is accurate as well as efficient 
 depends on the particular data set.

Similar to how a principal component analysis (PCA) 
can be used as a preprocessing step for other learning tasks,
in the next section we will use the tree tensor network algorithm 
to compute a reduced set of features for supervised learning. However, unlike
typical kernel PCA approaches, the tree tensor network approach discussed above imparts an explicit
layered structure to the resulting model. In future work it would be very interesting to 
analyze the layers to see if they learn a hierarchy of features as observed in works on 
deep neural networks \cite{Erhan:2009,Mehta:2014}. It may also be the case that the
learned tree tensor network representation of the data acts as a form of regularization when
used within other tasks.

\section{Supervised Optimization of the Top Tensor \label{sec:supervised}}

Having determined a tree like network $\mathcal{U}$ of isometry tensors,
we can now use this tree network as a starting point for optimizing 
a model for a supervised task. The specific model we explore in this section
is shown in Fig.~\ref{fig:treemodel},
and consists of the isometry layers $\mathcal{U}$ and a top tensor
$w$. We emphasize that the isometry layers forming the tree tensor network
$\mathcal{U}$ are obtained in an unsupervised
manner, just by using the criterion of projecting out directions in feature space not spanned
by the most significant eigenvectors of~$\rho$. 
Only the top tensor $w$ will be optimized for the supervised task
we are interested in; the layers $\mathcal{U}$ will be left fixed.
Of course we could optimize all of the layers for the supervised task,
but here we are interested in the question of whether layers computed
using a purely unsupervised algorithm are a good enough
representation of the data for other tasks.

Recall that when producing the layers $\mathcal{U}$, one progressively
coarse grains the training set feature vectors as shown in Fig.~\ref{fig:rg}.
Thus for the purposes of optimizing the top tensor $w$, one only needs to use the 
reduced representation of the training inputs $\mathbf{x}_j$ 
which are order-two tensors 
$\tilde{\Phi}^{t_1 t_2}(\mathbf{x}_j)$ defined as
\begin{align}
\tilde{\Phi}^{t_1 t_2}(\mathbf{x}_j) = 
\sum_{s_1,s_2,\ldots,s_N} \mathcal{U}^{t_1 t_2}_{s_1 s_2 \cdots s_N} \Phi^{s_1 s_2 \cdots s_N}(\mathbf{x}_j)
\ .
\end{align}
These coarse-grained feature tensors reside in the same space as $w_{t_1 t_2}$,
so we can conveniently write the supervised model as
\begin{align}
f(\mathbf{x}) = \sum_{t_1 t_2} w_{t_1 t_2} \tilde{\Phi}^{t_1 t_2}(\mathbf{x}) \ .
\end{align}
In the experiments below, we choose to define the supervised task using a quadratic cost 
function.

To extend the class of supervised models proposed above to the multi-task setting,
one generalizes the model $f(\mathbf{x})$ to a family of 
models $f^\ell(\mathbf{x})$ by training a collection of top tensors $w^\ell$, one
for each label $\ell$.
In this multi-task setting, we found it sufficient to compute a single tree network
 $\mathcal{U}$ which is shared between the different models (different top tensors) specialized 
for each task.

To test the proposal, we consider the MNIST dataset of grayscale images of handwritten digits, 
consisting of 60,000 training images and 10,000 test images of size \mbox{$28\times28$} with 
ten labels 0--9 \cite{MNIST}.
For the local feature maps $\phi^{s_n}(x_n)$ we choose
\begin{align}
\phi^{s_n=1}(x_n) &= 1 \nonumber \\
\phi^{s_n=2}(x_n) &= x_n  \label{eqn:local_map}
\end{align}
as proposed in Ref.~\onlinecite{Novikov:2016}. This is certainly not the only
choice of feature map one could make, but has the advantage of providing a simple
interpretation of the resulting model as a high-order polynomial.

To format the input, we scale each
pixel value $x_n$ to be in the interval $[0,1]$. To arrange the pixels of
the two-dimensional images into components $x_n$ of a vector, we rasterize the 
image, simply proceeding along the first row, then the second row, etc.
Other arrangements of the pixel data might lead to better outcomes and would
be interesting to explore, such as
choosing pixels which are spatially close in two dimensions to be grouped
together as components the vector $\mathbf{x}$.

The main hyper parameter controlling the experiments is the truncation error
cutoff $\epsilon$ used when making the tree isometry tensors.
To optimize the top tensor, we performed conjugate gradient
optimization until the cost function reached convergence, typically requiring a few hundred
iterations.
The bond dimensions of the isometry layers (dimensions of the internal indices between layers)
was very sensitive to the truncation error cutoff Eq.~(\ref{eqn:truncerr}) that was used, yet fortunately the performance of the model also
increased rapidly with lower cutoffs. 

For a truncation cutoff $\epsilon=10^{-3}$ we found the bond dimensions
connecting layers of isometries grew adaptively from about 3 between the first and second layers,
to a maximum of 15 between the sixth and seventh layers, finally leading to the top two
indices $t_1$ and $t_2$ having dimensions 107 and 151 respectively. Optimizing the top
tensor of this model for the supervised task gave a cost function value of $C=0.078$ 
yielding $98.75\%$ accuracy on the training set and $97.44\%$ accuracy on the test set.

For a cutoff $\epsilon=6\times10^{-4}$ we found the bond dimensions
connecting layers of isometries became gradually larger in intermediate layers, then 
rapidly increased in the topmost layers resulting in top indices $t_1$ and $t_2$
of size $328$ and $444$ respectively. Though this is perhaps a large coarse-grained feature
space in absolute terms, it represents a significant reduction (coarse graining) of the 
original feature space size of $2^{784}$. Optimizing the top tensor of this network 
for the supervised task gave a cost function value of $C=0.042$ 
yielding $99.68\%$ accuracy on the training set and $98.08\%$ accuracy on the test set.

Neither of the above experiments used an explicit form of regularization, such as
a weight penalty. Nevertheless the test set generalization was good. This suggests
that the form of the model based on training only the top layer for the specific task
while computing the lower layers just from statistical properties of the data could aid generalization. However, more work and experiments on more challenging data sets is needed to verify this intuition.


\section{Mixed Task-Specific / Unsupervised Algorithm  \label{sec:mixed}}

An interesting extension of the above approach is that one can mix the feature space covariance 
matrix $\rho$ defined in Eq.~(\ref{eqn:rho}) together with another matrix based on a specific task 
when computing the tree isometry tensors. 
In this way one can provide a prior guess for a supervised task, for example,
by adapting or biasing the tree tensors toward the prior guess for the task,
but still resulting in models with better performance than
the guess due to the
partially unsupervised nature of the algorithm. (Of course the specific task
could itself be unsupervised as well, such as fitting the distribution of the data
by minimizing the negative log likelihood.)

To provide an initial guess for the supervised problem, one can use a low-cost approach
such as a linear classifier $f_\text{lin}(\mathbf{x}) = V\cdot\,\mathbf{x}$.
Then a model of the form Eq.~(\ref{eqn:full_model}) can be 
written in terms of an MPS representation of the weights $W$ whose output is guaranteed
to be equal to that of the linear classifier. See Appendix~\ref{appendix:lin_mps} for
the details of this construction.
Of course the prior weight MPS does not have to be initialized by a linear classifier only, 
but could be further optimized for the supervised task using the techniques of Refs.~\onlinecite{Novikov:2016,Stoudenmire:2016s}.
By representing the initial guess as a tensor
network---such as a matrix product state (MPS)---one can retain the ability 
to efficiently compute the reduced covariance matrices needed for computing the tree tensors.

Assuming we have obtained a prior guess for supervised task weights in the form of an MPS,
begin the coarse graining procedure by defining
a covariance matrix $\rho_W$ from the provided weights as
\begin{align}
(\rho_W)^{\mathbf{s}}_{\mathbf{s}'} = W^{\dagger \mathbf{s}} W_{\mathbf{s}'} \ ,
\end{align}
or in the multi-task case,
\begin{align}
(\rho_W)^{\mathbf{s}}_{\mathbf{s}'} = \sum_\ell W^{\dagger \mathbf{s}}_\ell W^{\ell}_{\mathbf{s}'} \ .
\end{align}
Again define $\rho$ as a sum of outer products of the training data feature vectors as before
in Eq.~(\ref{eqn:rho}). Normalize both $\rho$ and $\rho_W$ to have unit trace, defining
\begin{align}
\hat{\rho}_W & = \frac{1}{\text{Tr}[\rho_W]}\, \rho_W \\
\hat{\rho}   & = \frac{1}{\text{Tr}[\rho]}\, \rho \ .
\end{align}

Now choose an empirical mixing parameter $\mu \in [0,1]$ and define $\rho_\mu$ as
\begin{align}
\rho_\mu = \mu\, \hat{\rho}_W + (1-\mu)\, \hat{\rho} 
\end{align}
so that for $\mu=0$ the covariance matrix $\rho_\mu$ will be the same as for the 
purely unsupervised case (up to normalization) but for $\mu > 0$ the tree tensors 
making up $\mathcal{U}$ will be adapted to  represent both the training data
and the weights $W$. For the case $\mu=1$, if one takes a small enough
cutoff when computing the tree tensors then the network $\mathcal{U}$ will be
adapted to exactly reproduce the provided weights $W$ regardless of the data.

\begin{figure}[t]
\includegraphics[width=0.95\columnwidth]{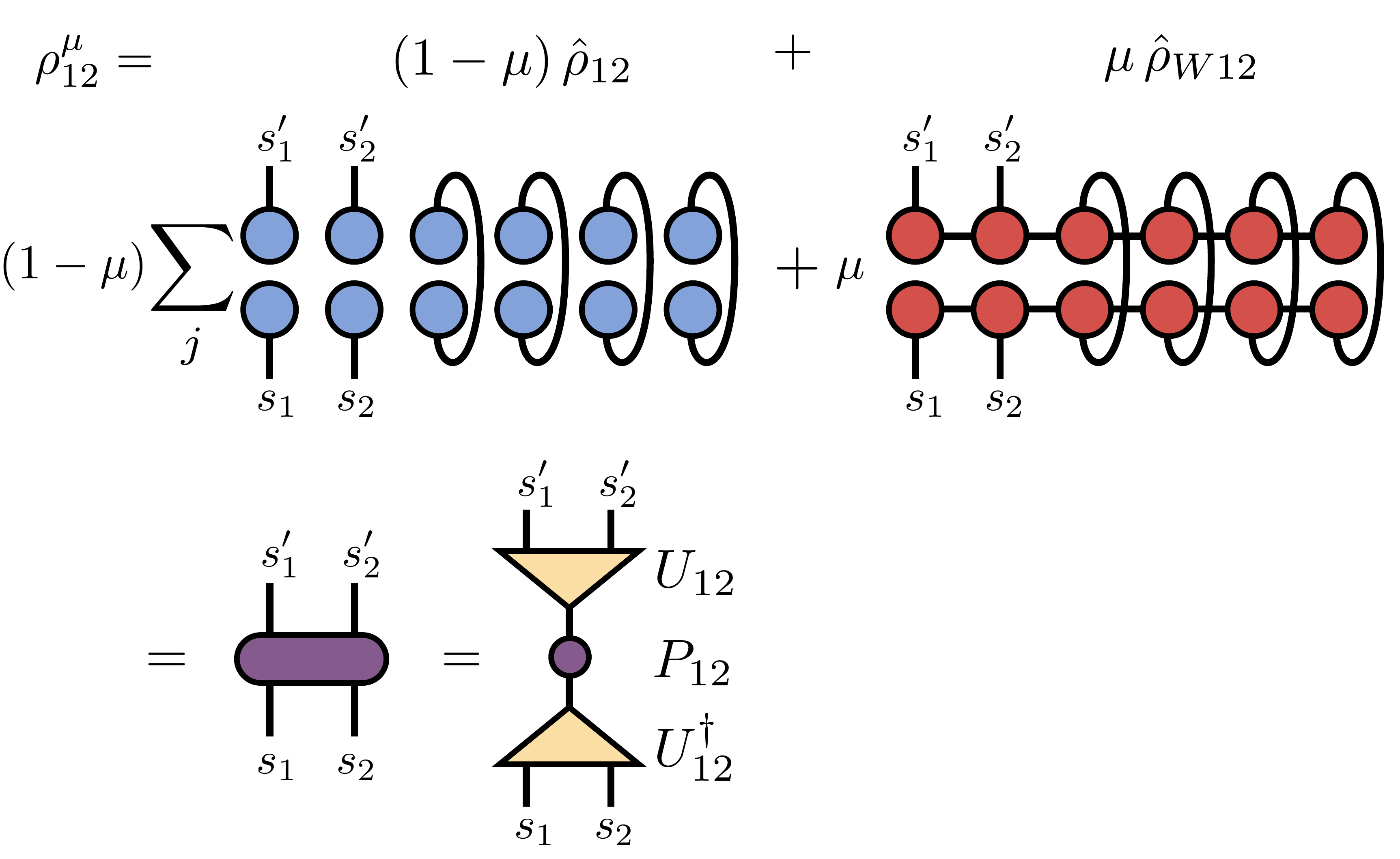}
\caption{In the mixed unsupervised/supervised algorithm for determining the tree 
tensors making up $\mathcal{U}$, the reduced covariance matrices are a weighted sum
of the reduced training data covariance matrix and reduced covariance matrix from
the provided supervised weights in MPS form. The figure above shows the computation
of the mixed reduced covariance for the first two sites; the computation for other
pairs of sites is similar just with different choices for which indices are traced
or left open.}
\label{fig:mixalg}
\end{figure}

Having defined $\rho_\mu$, the procedure to determine each layer of tree tensors
is quite similar to the purely unsupervised algorithm of 
Section~\ref{sec:unsupervised}, just with $\rho_\mu$ substituted 
for $\rho$. Figure \ref{fig:mixalg} shows the first step of the algorithm which is 
to compute the reduced covariance matrix $\rho_{\mu\, 12}$ from $\rho_\mu$, then 
diagonalize it to compute the first tree tensor.
 
The key algorithmic difference from the unsupervised case is that after determining
each layer, one must also coarse grain the provided weights $W$ along with the
training data so one can compute the reduced covariance matrices from $\rho_\mu$ 
at the next scale. Although the weights $W$ in MPS form have 
additional internal indices as shown in Fig.~\ref{fig:tns}(a), it is straightforward
to coarse grain an MPS with a tree tensor network layer: one simply contracts 
each isometry with pairs of MPS tensors.

For the case of a multi-class supervised task there will be multiple prior 
weight MPS $W^\ell$, one for each label $\ell$ 
(or one can equivalently provide a single MPS with 
an external or uncontracted label index). To generalize the above algorithm
to the multi-task setting, one defines the covariance matrix $\rho_W$ as
the sum over the covariance matrices of each of the prior supervised weights $W^\ell$
\begin{align}
(\rho_W)^{\mathbf{s}}_{\mathbf{s}'} = \sum_\ell W^{\dagger \mathbf{s}}_{\ell} W^{\ell}_{\mathbf{s}'} \ .
\end{align}

To test whether the strategy of mixing in a prior estimate of the supervised task weights
results in an improved model, we experiment again on the MNIST handwritten digits data set.
Using a mixing parameter $\mu=0.5$ and a truncation error cutoff $\epsilon=4\times10^{-4}$
results in a tree tensor network with top index sizes 279 and 393, where after
making the tree layers in a single pass, only the top tensor is  optimized further for
the supervised task. Despite the top index sizes
 being significantly smaller than those for the best experiment in Sec.~\ref{sec:supervised}
(where the sizes were 328 and 444), the results are slightly better: the cost function
value is $C=0.0325$, training set accuracy is $99.798\%$, and test set accuracy is $98.110\%$. 
This experiment strongly suggests that mixing weights trained for the supervised task
with the covariance matrix based purely on the data leads to a representation
of the data more suited for the specific task, which can be compressed further without
diminishing performance.

\begin{figure}[b]
\includegraphics[width=\columnwidth]{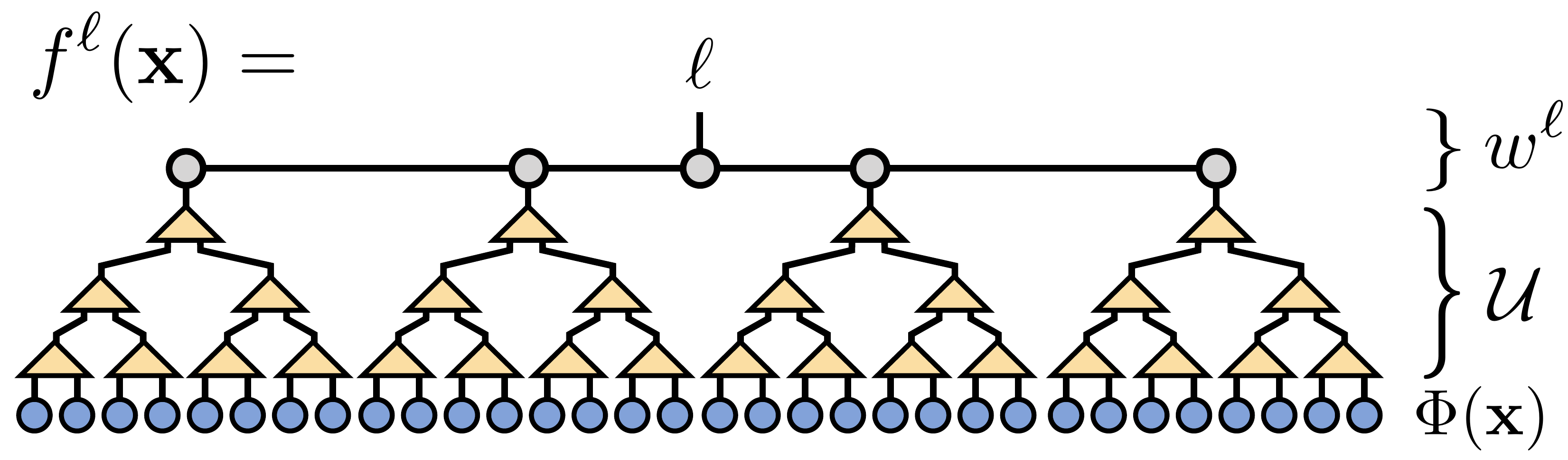}
\caption{Computing only a few tree layers results in a model with a high-order
top tensor. In the model above the top tensor $w^\ell$ is represented as a matrix product
state with a label index $\ell$ appropriate for the multi-task case.}
\label{fig:curtain}
\end{figure}

\section{Partial Coarse Graining: Tree Curtain Model \label{sec:partial}}

While the approaches in the previous sections involved computing tree tensor
networks with the maximum number of layers, computing fewer layers
can balance the benefits of a compressed data representation
against the loss of expressiveness from accumulated truncations
 when computing more layers.

One interesting aspect of computing fewer tree layers is that
after coarse graining, the data are still represented as high-order tensors, 
similar to Fig.~\ref{fig:rg}.
Specifically, the order of the data tensors after $R$ rescalings will be \mbox{$N_\text{top} = N/2^R$}.
Therefore to complete the model, the top tensor must also be a tensor of order $N_\text{top}$
if the output is to be a scalar, or order $N_\text{top}+1$ for vector-valued output in the multi-task
case. So to complete the model one can use another type of tensor network to represent the top tensor,
such as a matrix product state.

Choosing a matrix product state (MPS) form of the top tensor results in the architecture shown
in Fig.~\ref{fig:curtain}. After coarse graining the training data through the tree layers,
one can optimize the top MPS using previously developed methods for 
supervised \cite{Novikov:2016,Stoudenmire:2016s} or unsupervised \cite{Han:2017} learning tasks.
The resulting model resembles an MPS with a tree ``curtain'' attached.

\begin{figure}[t]
\includegraphics[width=0.4\columnwidth]{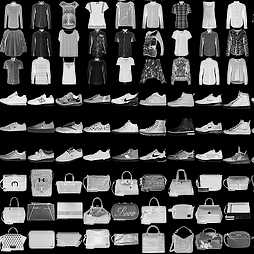}
\caption{Examples from the fashion MNIST data set \cite{Xiao:2017}.}
\label{fig:fashion}
\end{figure}

To test the effectiveness of the partial coarse-grained approach, and the resulting tree-curtain
model, we study the \emph{fashion} MNIST dataset \cite{Xiao:2017}. Similarly to MNIST, the data 
set consists of $28\times28$ grayscale images with ten labels, with 60,000 training and 10,000 test images.
However, the supervised learning task is significantly more challenging than MNIST because the 
images consist of photographs of a variety of clothing (shirts, shoes, etc.). Example
images from the data set are shown in Fig.~\ref{fig:fashion}.

To compute a tree curtain model for fashion MNIST, we first training a linear classifier 
resulting in only $83\%$ test accuracy. We then represent each of the linear classifier vectors $V^\ell$
as MPS and used the mixed covariance matrix approach (Sec.~\ref{sec:mixed}) with 
mixing parameter $\mu=0.9$ to optimize four tree tensor layers. Using
a truncation error cutoff $\epsilon=2\times10^{-9}$, the final coarse-grained feature 
indices attached to the top MPS reach a maximum of about 30. We then fix the 
internal bond dimension of the top MPS to be 300 and minimize the quadratic cost using 
thirty sweeps, or passes of alternating least squares (single-site DMRG) optimization. 

The optimized model reaches $95.38\%$ accuracy on the training set, and $88.97\%$ accuracy on the 
testing set. While $\sim\!\!89\%$ test accuracy is significantly less than achieved on the 
much easier MNIST  handwriting dataset, many of the available benchmarks using 
state-of-the-art approaches for fashion MNIST without preprocessing are in 
fact comparable to the results here, 
for example \mbox{XGBoost} ($89.8\%$), AlexNet ($89.9\%$),
and a two-layer convolutional neural network trained with Keras ($87.6\%$).
Better results are attainable; the best we are aware of is a GoogLeNet  
reaching $93.7\%$ test accuracy. But the fact that the architecture discussed
here yields similar performance to other powerful and standard approaches for a challenging
data set is an encouraging result. With further hyperparameter optimization and 
more efficient training algorithms such as stochastic gradient descent, we expect better
results can be achieved.

\section{Related Work}

The feature space covariance matrix \mbox{$\rho^\mathbf{s}_{\mathbf{s}'} = \sum_j \Phi^\mathbf{s}_j \Phi^{\dagger j}_{\mathbf{s}'}$} is closely 
related to the kernel  matrix \mbox{$K^{j}_{j'} = \sum_\mathbf{s} \Phi^{\dagger j}_{\mathbf{s}}  \Phi^\mathbf{s}_{j'}$}, which is a central quantity in the theory of kernel learning.
That both $\rho$ and $K$ have the same spectrum can be seen from the SVD of $\Phi^\mathbf{s}_j$.
Thus the idea of exploiting low-rank approximations to the kernel matrix 
is closely related to the present work \cite{Scholkopf:2001L,Bach:2013,Cesa-Bianchi}, 
as well as the idea of feature sampling
or random features \cite{Rahimi:2008,Rudi:2017}. But in contrast, the
the present work constructs a low-rank approximation for $\rho$  
directly in feature space, with an algorithm nevertheless scaling linearly in both training set
size and input space dimension. The tensor network form also imparts
an interesting structure onto the resulting model, allowing us to experiment with training
the top layer only. And we considered other choices besides $\rho$,
namely a mixed covariance matrix $\rho_\mu$, which produces a more compressed representation
of the data without reducing performance on a supervised task.

The algorithm in Sec.~\ref{sec:unsupervised} 
has many precedents in the tensor network literature, such as discussions by Vidal 
\cite{Vidal:2007,Vidal:2007a} of computing tree networks layer-wise to 
optimize fidelity of density matrices (pure and mixed states), and multigrid DMRG \cite{Dolfi:2012}.
Similar algorithms have also been employed for quantum state tomography \cite{Cramer:2010,Landon-Cardinal:2012}. Recent work by Nouy~\cite{Nouy:2017}
develops a related algorithm to reconstruct a function using a tree tensor network.

Another related line of work is the study of convolutional arithmetic 
circuits or ConvACs \cite{Cohen:2016,Cohen:2016c,Cohen:2016i}, 
a type of neural network constructed from a tree tensor network \cite{Levine:2017}. 
The present work could be viewed as a proposal for an 
unsupervised, adaptive algorithm for training \mbox{ConvAC} neural networks, along with variations
such as replacing the upper layers with an MPS. 

\section{Discussion}

The idea of coarse-graining in statistical physics motivates a similar 
approach to real-world data. The algorithm presented here resembles 
kernel PCA, but computed directly in feature space  
with the diagonalizing unitary approximated as a layered tensor network. 
Despite, or perhaps because of the approximations made, the resulting
coarse-grained representation of the data could be
used as a good starting point for other supervised or unsupervised tasks. 
The idea suggests interesting extensions such as exploring quantities 
besides the covariance matrix $\rho$ around which to base the algorithm. 
Using a mixed covariance matrix $\rho_\mu$  (Sec.~\ref{sec:mixed})
gave an improved representation, but there
could be other good choices. Instead of a tree tensor network, which has certain known
deficiencies as a coarse-graining scheme, it would be very interesting to use
a MERA tensor network \cite{Vidal:2007,Evenbly:2009}.

In an engineering sense, there is much room to improve the above algorithms, such as 
using tree tensors with two-dimensional groupings of indices \cite{Liu:2017}.
Another idea is feeding the optimized model back through the 
mixed algorithm of Sec.~\ref{sec:mixed} to further adapt
the tree tensors. 

That PCA is similar to coarse-graining, or the renormalization group,
has been observed recently by Bradde and Bialek \cite{Bradde:2016}. It would be very interesting to 
use the tools discussed here to analyze various data sets from a renormalization group perspective. 
An especially welcome outcome would be if the analysis could identify distinct
classes of data, and determine
which machine learning architectures have an inductive bias \cite{Cohen:2016i} suited to 
each class.

\section{Acknowledgements}

We thank David J. Schwab for an especially careful reading of the paper and many helpful comments. 
We also thank Glen Evenbly, William Huggins, Matthew Fishman, Steven White, John Terilla and Yiannis Vlassopoulos for helpful comments and discussions.

\appendix

\section{Tensor Diagram Notation \label{appendix:diagram}}

Tensor diagram notation is a simple, graphical notation for representing networks
of contractions of high-order tensors. The main rules of this notation are:
\begin{itemize}
\item A tensor with $n$ indices (an order $n$ tensor) is represented as a shape with
$n$ lines coming out of it. Each line represents a specific index but indices do not have
to be labeled or named when they can be distinguished by context.
\item Connecting index lines between a pair of tensors indicates that these indices are contracted, or summed over.
\end{itemize}
Other informal conventions may be used by specific authors, such as choosing special shapes or decorations
of tensors to denote special properties. For example, in this work a triangular shape indicates 
an isometric tensor. Unless otherwise specified, the only information that matters is the connectivity of the
network, with no fixed meaning given to the orientation or ordering of the index lines.
Furthermore, no special properties of the tensors are assumed such as symmetries or transformation
properties, unless specifically stated.

We emphasize that tensor diagram notation is completely rigorous, and just a way of notating
complicated sums. Unlike similar-looking neural network diagrams, every element of a tensor
diagram is a purely linear transformation (of the vector spaces whose basis elements are labeled by the  indices).

The main advantage of tensor diagram notation is that it frees one from having to assign names to
every index, which can be cumbersome for complicated tensor networks. 
The notation also makes it easier to compactly express sophisticated algorithms and to visually
understand different families of tensor networks. Various operations such as traces, outer products,
transposes, or matricization can be expressed implicitly in diagrammatic form without
requiring additional notation or symbols. For example, the diagrammatic expression of an outer 
product is simply the placement of two tensors nearby each other.

\section{Representing a Linear Classifier as an MPS \label{appendix:lin_mps}}

One can write a model of the 
form Eq.~(\ref{eqn:full_model}) with the weights in MPS form such that
the output of the model is guaranteed to be the same as a linear classifier
with weights $V$, assuming local feature maps of the form Eq.~(\ref{eqn:local_map})
namely $\phi(x) = [1,\ x]$.
This mapping was first discussed in Ref.~\onlinecite{Novikov:2016}.

Say we want to make our model output equivalent to a linear classifier
with parameters $V$
\begin{align}
f_\text{lin}(\mathbf{x}) = V\cdot\mathbf{x} = \sum_{n=1}^N V_j\, x^{j} \ \ .
\end{align}
First, assume the weights of our model Eq.~(\ref{eqn:full_model}) are in MPS form, meaning
\begin{align}
W^{s_1 s_2 s_3 \cdots s_N} = 
\sum_{\mathbf{\alpha}} A^{s_1}_{\alpha_1} A^{s_2}_{\alpha_2 \alpha_3}  A^{s_3}_{\alpha_3 \alpha_4} 
A^{s_N}_{\alpha_{N-1}}  \ . \label{eqn:mps}
\end{align}
Now define each MPS tensor $A^{s_j}_{\alpha_{j-1} \alpha_j}$ for each fixed value of 
$s_j$ as follows:
\begin{align}
A^{(s_j = 1)}_{\alpha_{j-1} \alpha_j} & = \begin{bmatrix}
1 & 0 \\
0 & 1
\end{bmatrix} \\
A^{(s_j = 2)}_{\alpha_{j-1} \alpha_j} & = \begin{bmatrix}
0 & 0 \\
V_j & 0
\end{bmatrix} \ .
\end{align}
For the first $A$ tensor one only takes the second row of each matrix and for the last $A$ tensor
one only takes the first column of each matrix. To include a constant shift, either add a 
fictitious input component $x^0 = 1$ to the definition of $\mathbf{x}$ or suitably adjust the first ``$A$'' tensor. This mapping from a linear classifier can also be extended to more general local 
feature maps $\phi^{s}(x)$ of dimension $d$ by suitably defining an ``extended linear classifier'' model
\begin{align}
f_\text{ext-lin.}(\mathbf{x}) = \sum_{n=1}^N \sum_{s=1}^d V_{ns} \phi^{s}(x_n)
\end{align}
and lifting this model to an MPS in a similar manner (the resulting MPS has bond dimension $d+1$).

\bibliography{semi}

\begin{thebibliography}{70}%
\makeatletter
\providecommand \@ifxundefined [1]{%
 \@ifx{#1\undefined}
}%
\providecommand \@ifnum [1]{%
 \ifnum #1\expandafter \@firstoftwo
 \else \expandafter \@secondoftwo
 \fi
}%
\providecommand \@ifx [1]{%
 \ifx #1\expandafter \@firstoftwo
 \else \expandafter \@secondoftwo
 \fi
}%
\providecommand \natexlab [1]{#1}%
\providecommand \enquote  [1]{``#1''}%
\providecommand \bibnamefont  [1]{#1}%
\providecommand \bibfnamefont [1]{#1}%
\providecommand \citenamefont [1]{#1}%
\providecommand \href@noop [0]{\@secondoftwo}%
\providecommand \href [0]{\begingroup \@sanitize@url \@href}%
\providecommand \@href[1]{\@@startlink{#1}\@@href}%
\providecommand \@@href[1]{\endgroup#1\@@endlink}%
\providecommand \@sanitize@url [0]{\catcode `\\12\catcode `\$12\catcode
  `\&12\catcode `\#12\catcode `\^12\catcode `\_12\catcode `\%12\relax}%
\providecommand \@@startlink[1]{}%
\providecommand \@@endlink[0]{}%
\providecommand \url  [0]{\begingroup\@sanitize@url \@url }%
\providecommand \@url [1]{\endgroup\@href {#1}{\urlprefix }}%
\providecommand \urlprefix  [0]{URL }%
\providecommand \Eprint [0]{\href }%
\providecommand \doibase [0]{http://dx.doi.org/}%
\providecommand \selectlanguage [0]{\@gobble}%
\providecommand \bibinfo  [0]{\@secondoftwo}%
\providecommand \bibfield  [0]{\@secondoftwo}%
\providecommand \translation [1]{[#1]}%
\providecommand \BibitemOpen [0]{}%
\providecommand \bibitemStop [0]{}%
\providecommand \bibitemNoStop [0]{.\EOS\space}%
\providecommand \EOS [0]{\spacefactor3000\relax}%
\providecommand \BibitemShut  [1]{\csname bibitem#1\endcsname}%
\let\auto@bib@innerbib\@empty
\bibitem [{\citenamefont {Anandkumar}\ \emph {et~al.}(2014)\citenamefont
  {Anandkumar}, \citenamefont {Ge}, \citenamefont {Hsu}, \citenamefont
  {Kakade},\ and\ \citenamefont {Telgarsky}}]{Anandkumar:2014}%
  \BibitemOpen
  \bibfield  {author} {\bibinfo {author} {\bibfnamefont {Animashree}\
  \bibnamefont {Anandkumar}}, \bibinfo {author} {\bibfnamefont {Rong}\
  \bibnamefont {Ge}}, \bibinfo {author} {\bibfnamefont {Daniel}\ \bibnamefont
  {Hsu}}, \bibinfo {author} {\bibfnamefont {Sham~M.}\ \bibnamefont {Kakade}}, \
  and\ \bibinfo {author} {\bibfnamefont {Matus}\ \bibnamefont {Telgarsky}},\
  }\bibfield  {title} {\enquote {\bibinfo {title} {Tensor decompositions for
  learning latent variable models},}\ }\href
  {http://jmlr.org/papers/v15/anandkumar14b.html} {\bibfield  {journal}
  {\bibinfo  {journal} {Journal of Machine Learning Research}\ }\textbf
  {\bibinfo {volume} {15}},\ \bibinfo {pages} {2773--2832} (\bibinfo {year}
  {2014})}\BibitemShut {NoStop}%
\bibitem [{\citenamefont {Sedghi}\ and\ \citenamefont
  {Anandkumar}(2016)}]{Sedghi:2016}%
  \BibitemOpen
  \bibfield  {author} {\bibinfo {author} {\bibfnamefont {Hanie}\ \bibnamefont
  {Sedghi}}\ and\ \bibinfo {author} {\bibfnamefont {Anima}\ \bibnamefont
  {Anandkumar}},\ }\bibfield  {title} {\enquote {\bibinfo {title} {Training
  input-output recurrent neural networks through spectral methods},}\
  }\href@noop {} {\bibfield  {journal} {\bibinfo  {journal} {arxiv:1603.00954}\
  } (\bibinfo {year} {2016})}\BibitemShut {NoStop}%
\bibitem [{\citenamefont {Novikov}\ \emph {et~al.}(2015)\citenamefont
  {Novikov}, \citenamefont {Podoprikhin}, \citenamefont {Osokin},\ and\
  \citenamefont {Vetrov}}]{Novikov:2015}%
  \BibitemOpen
  \bibfield  {author} {\bibinfo {author} {\bibfnamefont {Alexander}\
  \bibnamefont {Novikov}}, \bibinfo {author} {\bibfnamefont {Dmitry}\
  \bibnamefont {Podoprikhin}}, \bibinfo {author} {\bibfnamefont {Anton}\
  \bibnamefont {Osokin}}, \ and\ \bibinfo {author} {\bibfnamefont {Dmitry}\
  \bibnamefont {Vetrov}},\ }\bibfield  {title} {\enquote {\bibinfo {title}
  {Tensorizing neural networks},}\ }\href {http://arxiv.org/abs/1509.06569}
  {\bibfield  {journal} {\bibinfo  {journal} {arxiv:1509.06569}\ } (\bibinfo
  {year} {2015})}\BibitemShut {NoStop}%
\bibitem [{\citenamefont {Cohen}\ \emph {et~al.}(2016)\citenamefont {Cohen},
  \citenamefont {Sharir},\ and\ \citenamefont {Shashua}}]{Cohen:2016}%
  \BibitemOpen
  \bibfield  {author} {\bibinfo {author} {\bibfnamefont {Nadav}\ \bibnamefont
  {Cohen}}, \bibinfo {author} {\bibfnamefont {Or}~\bibnamefont {Sharir}}, \
  and\ \bibinfo {author} {\bibfnamefont {Amnon}\ \bibnamefont {Shashua}},\
  }\bibfield  {title} {\enquote {\bibinfo {title} {On the expressive power of
  deep learning: A tensor analysis},}\ }\href@noop {} {\bibfield  {journal}
  {\bibinfo  {journal} {29th Annual Conference on Learning Theory}\ ,\ \bibinfo
  {pages} {698--728}} (\bibinfo {year} {2016})}\BibitemShut {NoStop}%
\bibitem [{\citenamefont {Novikov}\ \emph {et~al.}()\citenamefont {Novikov},
  \citenamefont {Trofimov},\ and\ \citenamefont {Oseledets}}]{Novikov:2016}%
  \BibitemOpen
  \bibfield  {author} {\bibinfo {author} {\bibfnamefont {Alexander}\
  \bibnamefont {Novikov}}, \bibinfo {author} {\bibfnamefont {Mikhail}\
  \bibnamefont {Trofimov}}, \ and\ \bibinfo {author} {\bibfnamefont {Ivan}\
  \bibnamefont {Oseledets}},\ }\bibfield  {title} {\enquote {\bibinfo {title}
  {Exponential machines},}\ }\href {http://arxiv.org/abs/1605.03795} {\
  }\Eprint {http://arxiv.org/abs/arxiv:1605.03795 (2016)} {arxiv:1605.03795
  (2016)} \BibitemShut {NoStop}%
\bibitem [{\citenamefont {Stoudenmire}\ and\ \citenamefont
  {Schwab}(2016)}]{Stoudenmire:2016s}%
  \BibitemOpen
  \bibfield  {author} {\bibinfo {author} {\bibfnamefont {E.M.}\ \bibnamefont
  {Stoudenmire}}\ and\ \bibinfo {author} {\bibfnamefont {David~J}\ \bibnamefont
  {Schwab}},\ }\bibfield  {title} {\enquote {\bibinfo {title} {Supervised
  learning with tensor networks},}\ }in\ \href
  {http://papers.nips.cc/paper/6211-supervised-learning-with-tensor-networks.pdf}
  {\emph {\bibinfo {booktitle} {Advances In Neural Information Processing
  Systems 29}}}\ (\bibinfo {year} {2016})\ pp.\ \bibinfo {pages} {4799--4807},\
  \Eprint {http://arxiv.org/abs/arxiv:1605.05775} {arxiv:1605.05775}
  \BibitemShut {NoStop}%
\bibitem [{\citenamefont {Yu}\ \emph {et~al.}(2017)\citenamefont {Yu},
  \citenamefont {Zheng}, \citenamefont {Anandkumar},\ and\ \citenamefont
  {Yue}}]{Yu:2017}%
  \BibitemOpen
  \bibfield  {author} {\bibinfo {author} {\bibfnamefont {Rose}\ \bibnamefont
  {Yu}}, \bibinfo {author} {\bibfnamefont {Stephan}\ \bibnamefont {Zheng}},
  \bibinfo {author} {\bibfnamefont {Anima}\ \bibnamefont {Anandkumar}}, \ and\
  \bibinfo {author} {\bibfnamefont {Yisong}\ \bibnamefont {Yue}},\ }\bibfield
  {title} {\enquote {\bibinfo {title} {Long-term forecasting using tensor-train
  \mbox{RNNs}},}\ }\href@noop {} {\bibfield  {journal} {\bibinfo  {journal}
  {arxiv:1711.00073}\ } (\bibinfo {year} {2017})}\BibitemShut {NoStop}%
\bibitem [{\citenamefont {Kossaifi}\ \emph {et~al.}(2017)\citenamefont
  {Kossaifi}, \citenamefont {Lipton}, \citenamefont {Khanna}, \citenamefont
  {Furlanello},\ and\ \citenamefont {Anandkumar}}]{Kossaifi:2017}%
  \BibitemOpen
  \bibfield  {author} {\bibinfo {author} {\bibfnamefont {Jean}\ \bibnamefont
  {Kossaifi}}, \bibinfo {author} {\bibfnamefont {Zachary~C.}\ \bibnamefont
  {Lipton}}, \bibinfo {author} {\bibfnamefont {Aran}\ \bibnamefont {Khanna}},
  \bibinfo {author} {\bibfnamefont {Tommaso}\ \bibnamefont {Furlanello}}, \
  and\ \bibinfo {author} {\bibfnamefont {Anima}\ \bibnamefont {Anandkumar}},\
  }\bibfield  {title} {\enquote {\bibinfo {title} {Tensor contraction \&
  regression networks},}\ }\href@noop {} {\bibfield  {journal} {\bibinfo
  {journal} {arxiv:1707.08308}\ } (\bibinfo {year} {2017})}\BibitemShut
  {NoStop}%
\bibitem [{\citenamefont {Hallam}\ \emph {et~al.}(2017)\citenamefont {Hallam},
  \citenamefont {Grant}, \citenamefont {Stojevic}, \citenamefont {Severini},\
  and\ \citenamefont {Green}}]{Hallam:2017}%
  \BibitemOpen
  \bibfield  {author} {\bibinfo {author} {\bibfnamefont {Andrew}\ \bibnamefont
  {Hallam}}, \bibinfo {author} {\bibfnamefont {Edward}\ \bibnamefont {Grant}},
  \bibinfo {author} {\bibfnamefont {Vid}\ \bibnamefont {Stojevic}}, \bibinfo
  {author} {\bibfnamefont {Simone}\ \bibnamefont {Severini}}, \ and\ \bibinfo
  {author} {\bibfnamefont {Andrew~G.}\ \bibnamefont {Green}},\ }\bibfield
  {title} {\enquote {\bibinfo {title} {Compact neural networks based on the
  multiscale entanglement renormalization ansatz},}\ }\href@noop {} {\bibfield
  {journal} {\bibinfo  {journal} {arxiv:1711.03357}\ } (\bibinfo {year}
  {2017})}\BibitemShut {NoStop}%
\bibitem [{\citenamefont {Cohen}\ and\ \citenamefont
  {Shashua}(2016{\natexlab{a}})}]{Cohen:2016i}%
  \BibitemOpen
  \bibfield  {author} {\bibinfo {author} {\bibfnamefont {Nadav}\ \bibnamefont
  {Cohen}}\ and\ \bibinfo {author} {\bibfnamefont {Amnon}\ \bibnamefont
  {Shashua}},\ }\bibfield  {title} {\enquote {\bibinfo {title} {Inductive bias
  of deep convolutional networks through pooling geometry},}\ }\href@noop {}
  {\bibfield  {journal} {\bibinfo  {journal} {arxiv:1605.06743}\ } (\bibinfo
  {year} {2016}{\natexlab{a}})}\BibitemShut {NoStop}%
\bibitem [{\citenamefont {Blondel}\ \emph {et~al.}(2016)\citenamefont
  {Blondel}, \citenamefont {Ishihata}, \citenamefont {Fujino},\ and\
  \citenamefont {Ueda}}]{Blondel:2016}%
  \BibitemOpen
  \bibfield  {author} {\bibinfo {author} {\bibfnamefont {Mathieu}\ \bibnamefont
  {Blondel}}, \bibinfo {author} {\bibfnamefont {Masakazu}\ \bibnamefont
  {Ishihata}}, \bibinfo {author} {\bibfnamefont {Akinori}\ \bibnamefont
  {Fujino}}, \ and\ \bibinfo {author} {\bibfnamefont {Naonori}\ \bibnamefont
  {Ueda}},\ }\bibfield  {title} {\enquote {\bibinfo {title} {Polynomial
  networks and factorization machines: New insights and efficient training
  algorithms},}\ }\href@noop {} {\bibfield  {journal} {\bibinfo  {journal}
  {arxiv:1607.08810}\ } (\bibinfo {year} {2016})}\BibitemShut {NoStop}%
\bibitem [{\citenamefont {Orus}(2014)}]{Orus:2014a}%
  \BibitemOpen
  \bibfield  {author} {\bibinfo {author} {\bibfnamefont {Roman}\ \bibnamefont
  {Orus}},\ }\bibfield  {title} {\enquote {\bibinfo {title} {A practical
  introduction to tensor networks: Matrix product states and projected
  entangled pair states},}\ }\href@noop {} {\bibfield  {journal} {\bibinfo
  {journal} {Annals of Physics}\ }\textbf {\bibinfo {volume} {349}},\ \bibinfo
  {pages} {117 -- 158} (\bibinfo {year} {2014})}\BibitemShut {NoStop}%
\bibitem [{\citenamefont {Wilson}(1971)}]{Wilson:1971}%
  \BibitemOpen
  \bibfield  {author} {\bibinfo {author} {\bibfnamefont {Kenneth~G.}\
  \bibnamefont {Wilson}},\ }\bibfield  {title} {\enquote {\bibinfo {title}
  {Renormalization group and critical phenomena. i. renormalization group and
  the kadanoff scaling picture},}\ }\href@noop {} {\bibfield  {journal}
  {\bibinfo  {journal} {Phys. Rev. B}\ }\textbf {\bibinfo {volume} {4}},\
  \bibinfo {pages} {3174--3183} (\bibinfo {year} {1971})}\BibitemShut {NoStop}%
\bibitem [{\citenamefont {Wilson}(1979)}]{Wilson:1979}%
  \BibitemOpen
  \bibfield  {author} {\bibinfo {author} {\bibfnamefont {Kenneth~G.}\
  \bibnamefont {Wilson}},\ }\bibfield  {title} {\enquote {\bibinfo {title}
  {Problems in physics with many scales of length},}\ }\href@noop {} {\bibfield
   {journal} {\bibinfo  {journal} {Scientific American}\ ,\ \bibinfo {pages}
  {158--179}} (\bibinfo {year} {1979})}\BibitemShut {NoStop}%
\bibitem [{\citenamefont {Evenbly}\ and\ \citenamefont
  {White}(2016)}]{Evenbly:2016w}%
  \BibitemOpen
  \bibfield  {author} {\bibinfo {author} {\bibfnamefont {Glen}\ \bibnamefont
  {Evenbly}}\ and\ \bibinfo {author} {\bibfnamefont {Steven~R.}\ \bibnamefont
  {White}},\ }\bibfield  {title} {\enquote {\bibinfo {title} {Representation
  and design of wavelets using unitary circuits},}\ }\href@noop {} {\bibfield
  {journal} {\bibinfo  {journal} {arxiv:1605.07312}\ } (\bibinfo {year}
  {2016})}\BibitemShut {NoStop}%
\bibitem [{\citenamefont {Erhan}\ \emph {et~al.}(2009)\citenamefont {Erhan},
  \citenamefont {Bengio}, \citenamefont {Courville},\ and\ \citenamefont
  {Vincent}}]{Erhan:2009}%
  \BibitemOpen
  \bibfield  {author} {\bibinfo {author} {\bibfnamefont {Dumitru}\ \bibnamefont
  {Erhan}}, \bibinfo {author} {\bibfnamefont {Yoshua}\ \bibnamefont {Bengio}},
  \bibinfo {author} {\bibfnamefont {Aaron}\ \bibnamefont {Courville}}, \ and\
  \bibinfo {author} {\bibfnamefont {Pascal}\ \bibnamefont {Vincent}},\
  }\bibfield  {title} {\enquote {\bibinfo {title} {Visualizing higher-layer
  features of a deep network},}\ }\href@noop {} {\bibfield  {journal} {\bibinfo
   {journal} {University of Montreal}\ } (\bibinfo {year} {2009})}\BibitemShut
  {NoStop}%
\bibitem [{\citenamefont {Mehta}\ and\ \citenamefont
  {Schwab}(2014)}]{Mehta:2014}%
  \BibitemOpen
  \bibfield  {author} {\bibinfo {author} {\bibfnamefont {Pankaj}\ \bibnamefont
  {Mehta}}\ and\ \bibinfo {author} {\bibfnamefont {David}\ \bibnamefont
  {Schwab}},\ }\bibfield  {title} {\enquote {\bibinfo {title} {An exact mapping
  between the variational renormalization group and deep learning},}\
  }\href@noop {} {\bibfield  {journal} {\bibinfo  {journal} {arxiv:1410.3831}\
  } (\bibinfo {year} {2014})}\BibitemShut {NoStop}%
\bibitem [{\citenamefont {Bradde}\ and\ \citenamefont
  {Bialek}(2016)}]{Bradde:2016}%
  \BibitemOpen
  \bibfield  {author} {\bibinfo {author} {\bibfnamefont {Serena}\ \bibnamefont
  {Bradde}}\ and\ \bibinfo {author} {\bibfnamefont {William}\ \bibnamefont
  {Bialek}},\ }\bibfield  {title} {\enquote {\bibinfo {title} {\mbox{PCA} meets
  \mbox{RG}},}\ }\href@noop {} {\bibfield  {journal} {\bibinfo  {journal}
  {arxiv:1610.09733}\ } (\bibinfo {year} {2016})}\BibitemShut {NoStop}%
\bibitem [{\citenamefont {Fannes}\ \emph {et~al.}(1992)\citenamefont {Fannes},
  \citenamefont {Nachtergaele},\ and\ \citenamefont {Werner}}]{Fannes:1992}%
  \BibitemOpen
  \bibfield  {author} {\bibinfo {author} {\bibfnamefont {M.}~\bibnamefont
  {Fannes}}, \bibinfo {author} {\bibfnamefont {B.}~\bibnamefont
  {Nachtergaele}}, \ and\ \bibinfo {author} {\bibfnamefont {R.~F.}\
  \bibnamefont {Werner}},\ }\bibfield  {title} {\enquote {\bibinfo {title}
  {Finitely correlated states on quantum spin chains},}\ }\href@noop {}
  {\bibfield  {journal} {\bibinfo  {journal} {Communications in Mathematical
  Physics}\ }\textbf {\bibinfo {volume} {144}},\ \bibinfo {pages} {443--490}
  (\bibinfo {year} {1992})}\BibitemShut {NoStop}%
\bibitem [{\citenamefont {\"Ostlund}\ and\ \citenamefont
  {Rommer}(1995)}]{Ostlund:1995}%
  \BibitemOpen
  \bibfield  {author} {\bibinfo {author} {\bibfnamefont {Stellan}\ \bibnamefont
  {\"Ostlund}}\ and\ \bibinfo {author} {\bibfnamefont {Stefan}\ \bibnamefont
  {Rommer}},\ }\bibfield  {title} {\enquote {\bibinfo {title} {Thermodynamic
  limit of density matrix renormalization},}\ }\href@noop {} {\bibfield
  {journal} {\bibinfo  {journal} {Phys. Rev. Lett.}\ }\textbf {\bibinfo
  {volume} {75}},\ \bibinfo {pages} {3537--3540} (\bibinfo {year}
  {1995})}\BibitemShut {NoStop}%
\bibitem [{\citenamefont {Vidal}(2003)}]{Vidal:2003}%
  \BibitemOpen
  \bibfield  {author} {\bibinfo {author} {\bibfnamefont {Guifr\'e}\
  \bibnamefont {Vidal}},\ }\bibfield  {title} {\enquote {\bibinfo {title}
  {Efficient classical simulation of slightly entangled quantum
  computations},}\ }\href@noop {} {\bibfield  {journal} {\bibinfo  {journal}
  {Phys. Rev. Lett.}\ }\textbf {\bibinfo {volume} {91}},\ \bibinfo {pages}
  {147902} (\bibinfo {year} {2003})}\BibitemShut {NoStop}%
\bibitem [{\citenamefont {Verstraete}\ and\ \citenamefont
  {Cirac}(2004)}]{Verstraete:2004p}%
  \BibitemOpen
  \bibfield  {author} {\bibinfo {author} {\bibfnamefont {F.}~\bibnamefont
  {Verstraete}}\ and\ \bibinfo {author} {\bibfnamefont {J.~I.}\ \bibnamefont
  {Cirac}},\ }\bibfield  {title} {\enquote {\bibinfo {title} {Renormalization
  algorithms for quantum-many body systems in two and higher dimensions},}\
  }\href@noop {} {\bibfield  {journal} {\bibinfo  {journal} {cond-mat/0407066}\
  } (\bibinfo {year} {2004})}\BibitemShut {NoStop}%
\bibitem [{\citenamefont {Vidal}(2007{\natexlab{a}})}]{Vidal:2007}%
  \BibitemOpen
  \bibfield  {author} {\bibinfo {author} {\bibfnamefont {G.}~\bibnamefont
  {Vidal}},\ }\bibfield  {title} {\enquote {\bibinfo {title} {Entanglement
  renormalization},}\ }\href@noop {} {\bibfield  {journal} {\bibinfo  {journal}
  {Phys. Rev. Lett.}\ }\textbf {\bibinfo {volume} {99}},\ \bibinfo {pages}
  {220405} (\bibinfo {year} {2007}{\natexlab{a}})}\BibitemShut {NoStop}%
\bibitem [{\citenamefont {White}(1992)}]{White:1992}%
  \BibitemOpen
  \bibfield  {author} {\bibinfo {author} {\bibfnamefont {Steven~R.}\
  \bibnamefont {White}},\ }\bibfield  {title} {\enquote {\bibinfo {title}
  {Density matrix formulation for quantum renormalization groups},}\
  }\href@noop {} {\bibfield  {journal} {\bibinfo  {journal} {Phys. Rev. Lett.}\
  }\textbf {\bibinfo {volume} {69}},\ \bibinfo {pages} {2863--2866} (\bibinfo
  {year} {1992})}\BibitemShut {NoStop}%
\bibitem [{\citenamefont {Schollw{\"o}ck}(2005)}]{Schollwoeck:2005}%
  \BibitemOpen
  \bibfield  {author} {\bibinfo {author} {\bibfnamefont {U.}~\bibnamefont
  {Schollw{\"o}ck}},\ }\bibfield  {title} {\enquote {\bibinfo {title} {The
  density-matrix renormalization group},}\ }\href@noop {} {\bibfield  {journal}
  {\bibinfo  {journal} {Rev. Mod. Phys.}\ }\textbf {\bibinfo {volume} {77}},\
  \bibinfo {pages} {259--315} (\bibinfo {year} {2005})}\BibitemShut {NoStop}%
\bibitem [{\citenamefont {McCulloch}(2007)}]{McCulloch:2007}%
  \BibitemOpen
  \bibfield  {author} {\bibinfo {author} {\bibfnamefont {Ian~P.}\ \bibnamefont
  {McCulloch}},\ }\bibfield  {title} {\enquote {\bibinfo {title} {From
  density-matrix renormalization group to matrix product states},}\ }\href@noop
  {} {\bibfield  {journal} {\bibinfo  {journal} {J. Stat. Mech.}\ ,\ \bibinfo
  {pages} {P10014}} (\bibinfo {year} {2007})}\BibitemShut {NoStop}%
\bibitem [{\citenamefont {McCulloch}(2008)}]{McCulloch:2008}%
  \BibitemOpen
  \bibfield  {author} {\bibinfo {author} {\bibfnamefont {I.~P.}\ \bibnamefont
  {McCulloch}},\ }\bibfield  {title} {\enquote {\bibinfo {title} {Infinite size
  density matrix renormalization group, revisited},}\ }\href@noop {} {\bibfield
   {journal} {\bibinfo  {journal} {arxiv:0804.2509}\ } (\bibinfo {year}
  {2008})}\BibitemShut {NoStop}%
\bibitem [{\citenamefont {Schollw\"ock}(2011)}]{Schollwoeck:2011}%
  \BibitemOpen
  \bibfield  {author} {\bibinfo {author} {\bibfnamefont {U.}~\bibnamefont
  {Schollw\"ock}},\ }\bibfield  {title} {\enquote {\bibinfo {title} {The
  density-matrix renormalization group in the age of matrix product states},}\
  }\href@noop {} {\bibfield  {journal} {\bibinfo  {journal} {Annals of
  Physics}\ }\textbf {\bibinfo {volume} {326}},\ \bibinfo {pages} {96--192}
  (\bibinfo {year} {2011})}\BibitemShut {NoStop}%
\bibitem [{\citenamefont {Levin}\ and\ \citenamefont
  {Nave}(2007)}]{Levin:2007}%
  \BibitemOpen
  \bibfield  {author} {\bibinfo {author} {\bibfnamefont {Michael}\ \bibnamefont
  {Levin}}\ and\ \bibinfo {author} {\bibfnamefont {Cody~P.}\ \bibnamefont
  {Nave}},\ }\bibfield  {title} {\enquote {\bibinfo {title} {Tensor
  renormalization group approach to two-dimensional classical lattice
  models},}\ }\href@noop {} {\bibfield  {journal} {\bibinfo  {journal} {Phys.
  Rev. Lett.}\ }\textbf {\bibinfo {volume} {99}},\ \bibinfo {pages} {120601}
  (\bibinfo {year} {2007})}\BibitemShut {NoStop}%
\bibitem [{\citenamefont {Evenbly}\ and\ \citenamefont
  {Vidal}(2015)}]{Evenbly:2014}%
  \BibitemOpen
  \bibfield  {author} {\bibinfo {author} {\bibfnamefont {G.}~\bibnamefont
  {Evenbly}}\ and\ \bibinfo {author} {\bibfnamefont {G.}~\bibnamefont
  {Vidal}},\ }\bibfield  {title} {\enquote {\bibinfo {title} {Tensor network
  renormalization},}\ }\href {\doibase 10.1103/PhysRevLett.115.180405}
  {\bibfield  {journal} {\bibinfo  {journal} {Phys. Rev. Lett.}\ }\textbf
  {\bibinfo {volume} {115}},\ \bibinfo {pages} {180405} (\bibinfo {year}
  {2015})}\BibitemShut {NoStop}%
\bibitem [{\citenamefont {Evenbly}\ and\ \citenamefont
  {Vidal}(2009)}]{Evenbly:2009}%
  \BibitemOpen
  \bibfield  {author} {\bibinfo {author} {\bibfnamefont {G.}~\bibnamefont
  {Evenbly}}\ and\ \bibinfo {author} {\bibfnamefont {G.}~\bibnamefont
  {Vidal}},\ }\bibfield  {title} {\enquote {\bibinfo {title} {Algorithms for
  entanglement renormalization},}\ }\href {\doibase 10.1103/PhysRevB.79.144108}
  {\bibfield  {journal} {\bibinfo  {journal} {Phys. Rev. B}\ }\textbf {\bibinfo
  {volume} {79}},\ \bibinfo {pages} {144108} (\bibinfo {year}
  {2009})}\BibitemShut {NoStop}%
\bibitem [{\citenamefont {Evenbly}\ and\ \citenamefont
  {Vidal}(2011{\natexlab{a}})}]{Evenbly:2011g}%
  \BibitemOpen
  \bibfield  {author} {\bibinfo {author} {\bibfnamefont {G.}~\bibnamefont
  {Evenbly}}\ and\ \bibinfo {author} {\bibfnamefont {G.}~\bibnamefont
  {Vidal}},\ }\bibfield  {title} {\enquote {\bibinfo {title} {Tensor network
  states and geometry},}\ }\href@noop {} {\bibfield  {journal} {\bibinfo
  {journal} {Journal of Statistical Physics}\ }\textbf {\bibinfo {volume}
  {145}},\ \bibinfo {pages} {891--918} (\bibinfo {year}
  {2011}{\natexlab{a}})}\BibitemShut {NoStop}%
\bibitem [{\citenamefont {Swingle}(2012)}]{Swingle:2012m}%
  \BibitemOpen
  \bibfield  {author} {\bibinfo {author} {\bibfnamefont {Brian}\ \bibnamefont
  {Swingle}},\ }\bibfield  {title} {\enquote {\bibinfo {title} {Entanglement
  renormalization and holography},}\ }\href@noop {} {\bibfield  {journal}
  {\bibinfo  {journal} {Phys. Rev. D}\ }\textbf {\bibinfo {volume} {86}},\
  \bibinfo {pages} {065007} (\bibinfo {year} {2012})}\BibitemShut {NoStop}%
\bibitem [{\citenamefont {Nozaki}\ \emph {et~al.}(2012)\citenamefont {Nozaki},
  \citenamefont {Ryu},\ and\ \citenamefont {Takayanagi}}]{Nozaki:2012}%
  \BibitemOpen
  \bibfield  {author} {\bibinfo {author} {\bibfnamefont {Masahiro}\
  \bibnamefont {Nozaki}}, \bibinfo {author} {\bibfnamefont {Shinsei}\
  \bibnamefont {Ryu}}, \ and\ \bibinfo {author} {\bibfnamefont {Tadashi}\
  \bibnamefont {Takayanagi}},\ }\bibfield  {title} {\enquote {\bibinfo {title}
  {Holographic geometry of entanglement renormalization in quantum field
  theories},}\ }\href {\doibase 10.1007/JHEP10(2012)193} {\bibfield  {journal}
  {\bibinfo  {journal} {Journal of High Energy Physics}\ }\textbf {\bibinfo
  {volume} {2012}},\ \bibinfo {pages} {193} (\bibinfo {year}
  {2012})}\BibitemShut {NoStop}%
\bibitem [{\citenamefont {Hayden}\ \emph {et~al.}(2016)\citenamefont {Hayden},
  \citenamefont {Nezami}, \citenamefont {Qi}, \citenamefont {Thomas},
  \citenamefont {Walter},\ and\ \citenamefont {Yang}}]{Hayden:2016}%
  \BibitemOpen
  \bibfield  {author} {\bibinfo {author} {\bibfnamefont {Patrick}\ \bibnamefont
  {Hayden}}, \bibinfo {author} {\bibfnamefont {Sepehr}\ \bibnamefont {Nezami}},
  \bibinfo {author} {\bibfnamefont {Xiao-Liang}\ \bibnamefont {Qi}}, \bibinfo
  {author} {\bibfnamefont {Nathaniel}\ \bibnamefont {Thomas}}, \bibinfo
  {author} {\bibfnamefont {Michael}\ \bibnamefont {Walter}}, \ and\ \bibinfo
  {author} {\bibfnamefont {Zhao}\ \bibnamefont {Yang}},\ }\bibfield  {title}
  {\enquote {\bibinfo {title} {Holographic duality from random tensor
  networks},}\ }\href@noop {} {\bibfield  {journal} {\bibinfo  {journal}
  {Journal of High Energy Physics}\ }\textbf {\bibinfo {volume} {2016}},\
  \bibinfo {pages} {9} (\bibinfo {year} {2016})}\BibitemShut {NoStop}%
\bibitem [{\citenamefont {Levine}\ \emph {et~al.}(2017)\citenamefont {Levine},
  \citenamefont {Yakira}, \citenamefont {Cohen},\ and\ \citenamefont
  {Shashua}}]{Levine:2017}%
  \BibitemOpen
  \bibfield  {author} {\bibinfo {author} {\bibfnamefont {Yoav}\ \bibnamefont
  {Levine}}, \bibinfo {author} {\bibfnamefont {David}\ \bibnamefont {Yakira}},
  \bibinfo {author} {\bibfnamefont {Nadav}\ \bibnamefont {Cohen}}, \ and\
  \bibinfo {author} {\bibfnamefont {Amnon}\ \bibnamefont {Shashua}},\
  }\bibfield  {title} {\enquote {\bibinfo {title} {Deep learning and quantum
  entanglement: Fundamental connections with implications to network design},}\
  }\href {https://arxiv.org/abs/1704.01552} {\  (\bibinfo {year} {2017})},\
  \Eprint {http://arxiv.org/abs/arxiv:1704.01552} {arxiv:1704.01552}
  \BibitemShut {NoStop}%
\bibitem [{\citenamefont {Liu}\ \emph {et~al.}(2017)\citenamefont {Liu},
  \citenamefont {Ran}, \citenamefont {Wittek}, \citenamefont {Peng},
  \citenamefont {Garc{\'\i}a}, \citenamefont {Su},\ and\ \citenamefont
  {Lewenstein}}]{Liu:2017}%
  \BibitemOpen
  \bibfield  {author} {\bibinfo {author} {\bibfnamefont {Ding}\ \bibnamefont
  {Liu}}, \bibinfo {author} {\bibfnamefont {Shi-Ju}\ \bibnamefont {Ran}},
  \bibinfo {author} {\bibfnamefont {Peter}\ \bibnamefont {Wittek}}, \bibinfo
  {author} {\bibfnamefont {Cheng}\ \bibnamefont {Peng}}, \bibinfo {author}
  {\bibfnamefont {Raul~Bl{\'a}zquez}\ \bibnamefont {Garc{\'\i}a}}, \bibinfo
  {author} {\bibfnamefont {Gang}\ \bibnamefont {Su}}, \ and\ \bibinfo {author}
  {\bibfnamefont {Maciej}\ \bibnamefont {Lewenstein}},\ }\bibfield  {title}
  {\enquote {\bibinfo {title} {Machine learning by two-dimensional hierarchical
  tensor networks: A quantum information theoretic perspective on deep
  architectures},}\ }\href@noop {} {\bibfield  {journal} {\bibinfo  {journal}
  {arxiv:1710.04833}\ } (\bibinfo {year} {2017})}\BibitemShut {NoStop}%
\bibitem [{\citenamefont {Khrulkov}\ \emph {et~al.}(2017)\citenamefont
  {Khrulkov}, \citenamefont {Novikov},\ and\ \citenamefont
  {Oseledets}}]{Khrulkov:2017}%
  \BibitemOpen
  \bibfield  {author} {\bibinfo {author} {\bibfnamefont {Valentin}\
  \bibnamefont {Khrulkov}}, \bibinfo {author} {\bibfnamefont {Alexander}\
  \bibnamefont {Novikov}}, \ and\ \bibinfo {author} {\bibfnamefont {Ivan}\
  \bibnamefont {Oseledets}},\ }\bibfield  {title} {\enquote {\bibinfo {title}
  {Expressive power of recurrent neural networks},}\ }\href@noop {} {\bibfield
  {journal} {\bibinfo  {journal} {arxiv:1711.00811}\ } (\bibinfo {year}
  {2017})}\BibitemShut {NoStop}%
\bibitem [{\citenamefont {Han}\ \emph {et~al.}(2017)\citenamefont {Han},
  \citenamefont {Wang}, \citenamefont {Fan}, \citenamefont {Wang},\ and\
  \citenamefont {Zhang}}]{Han:2017}%
  \BibitemOpen
  \bibfield  {author} {\bibinfo {author} {\bibfnamefont {Zhao-Yu}\ \bibnamefont
  {Han}}, \bibinfo {author} {\bibfnamefont {Jun}\ \bibnamefont {Wang}},
  \bibinfo {author} {\bibfnamefont {Heng}\ \bibnamefont {Fan}}, \bibinfo
  {author} {\bibfnamefont {Lei}\ \bibnamefont {Wang}}, \ and\ \bibinfo {author}
  {\bibfnamefont {Pan}\ \bibnamefont {Zhang}},\ }\bibfield  {title} {\enquote
  {\bibinfo {title} {Unsupervised generative modeling using matrix product
  states},}\ }\href@noop {} {\bibfield  {journal} {\bibinfo  {journal}
  {arxiv:1709.01662}\ } (\bibinfo {year} {2017})}\BibitemShut {NoStop}%
\bibitem [{\citenamefont {Cohen}\ and\ \citenamefont
  {Shashua}(2016{\natexlab{b}})}]{Cohen:2016c}%
  \BibitemOpen
  \bibfield  {author} {\bibinfo {author} {\bibfnamefont {Nadav}\ \bibnamefont
  {Cohen}}\ and\ \bibinfo {author} {\bibfnamefont {Amnon}\ \bibnamefont
  {Shashua}},\ }\bibfield  {title} {\enquote {\bibinfo {title} {Convolutional
  rectifier networks as generalized tensor decompositions},}\ }\href@noop {}
  {\bibfield  {journal} {\bibinfo  {journal} {arxiv:1603.00162}\ } (\bibinfo
  {year} {2016}{\natexlab{b}})}\BibitemShut {NoStop}%
\bibitem [{\citenamefont {Perez-Garcia}\ \emph {et~al.}(2007)\citenamefont
  {Perez-Garcia}, \citenamefont {Verstraete}, \citenamefont {Wolf},\ and\
  \citenamefont {Cirac}}]{Perez-Garcia:2007}%
  \BibitemOpen
  \bibfield  {author} {\bibinfo {author} {\bibfnamefont {D.}~\bibnamefont
  {Perez-Garcia}}, \bibinfo {author} {\bibfnamefont {F.}~\bibnamefont
  {Verstraete}}, \bibinfo {author} {\bibfnamefont {M.~M.}\ \bibnamefont
  {Wolf}}, \ and\ \bibinfo {author} {\bibfnamefont {J.~I.}\ \bibnamefont
  {Cirac}},\ }\bibfield  {title} {\enquote {\bibinfo {title} {Matrix product
  state representations},}\ }\href@noop {} {\bibfield  {journal} {\bibinfo
  {journal} {Quantum Info. Comput.}\ }\textbf {\bibinfo {volume} {7}},\
  \bibinfo {pages} {401--430} (\bibinfo {year} {2007})}\BibitemShut {NoStop}%
\bibitem [{\citenamefont {Schuch}\ \emph {et~al.}(2010)\citenamefont {Schuch},
  \citenamefont {Cirac},\ and\ \citenamefont
  {P{\'e}rez-Garc{\'\i}a}}]{Schuch:2010p}%
  \BibitemOpen
  \bibfield  {author} {\bibinfo {author} {\bibfnamefont {Norbert}\ \bibnamefont
  {Schuch}}, \bibinfo {author} {\bibfnamefont {Ignacio}\ \bibnamefont {Cirac}},
  \ and\ \bibinfo {author} {\bibfnamefont {David}\ \bibnamefont
  {P{\'e}rez-Garc{\'\i}a}},\ }\bibfield  {title} {\enquote {\bibinfo {title}
  {Peps as ground states: Degeneracy and topology},}\ }\href@noop {} {\bibfield
   {journal} {\bibinfo  {journal} {Annals of Physics}\ }\textbf {\bibinfo
  {volume} {325}},\ \bibinfo {pages} {2153 -- 2192} (\bibinfo {year}
  {2010})}\BibitemShut {NoStop}%
\bibitem [{\citenamefont {Or\'us}\ and\ \citenamefont
  {Vidal}(2008)}]{Orus:2008}%
  \BibitemOpen
  \bibfield  {author} {\bibinfo {author} {\bibfnamefont {R.}~\bibnamefont
  {Or\'us}}\ and\ \bibinfo {author} {\bibfnamefont {G.}~\bibnamefont {Vidal}},\
  }\bibfield  {title} {\enquote {\bibinfo {title} {Infinite time-evolving block
  decimation algorithm beyond unitary evolution},}\ }\href@noop {} {\bibfield
  {journal} {\bibinfo  {journal} {Phys. Rev. B}\ }\textbf {\bibinfo {volume}
  {78}},\ \bibinfo {pages} {155117} (\bibinfo {year} {2008})}\BibitemShut
  {NoStop}%
\bibitem [{\citenamefont {Chan}(2008)}]{Chan:2008a}%
  \BibitemOpen
  \bibfield  {author} {\bibinfo {author} {\bibfnamefont {Garnet Kin-Lic}\
  \bibnamefont {Chan}},\ }\bibfield  {title} {\enquote {\bibinfo {title}
  {Density matrix renormalization group lagrangians},}\ }\href@noop {}
  {\bibfield  {journal} {\bibinfo  {journal} {arxiv:0804.1755}\ } (\bibinfo
  {year} {2008})}\BibitemShut {NoStop}%
\bibitem [{\citenamefont {Ferris}\ and\ \citenamefont
  {Vidal}(2012)}]{Ferris:2012}%
  \BibitemOpen
  \bibfield  {author} {\bibinfo {author} {\bibfnamefont {Andrew~J.}\
  \bibnamefont {Ferris}}\ and\ \bibinfo {author} {\bibfnamefont {Guifre}\
  \bibnamefont {Vidal}},\ }\bibfield  {title} {\enquote {\bibinfo {title}
  {Perfect sampling with unitary tensor networks},}\ }\href@noop {} {\bibfield
  {journal} {\bibinfo  {journal} {Phys. Rev. B}\ }\textbf {\bibinfo {volume}
  {85}},\ \bibinfo {pages} {165146} (\bibinfo {year} {2012})}\BibitemShut
  {NoStop}%
\bibitem [{\citenamefont {Stoudenmire}\ and\ \citenamefont
  {White}(2010)}]{Stoudenmire:2010}%
  \BibitemOpen
  \bibfield  {author} {\bibinfo {author} {\bibfnamefont {E.~M.}\ \bibnamefont
  {Stoudenmire}}\ and\ \bibinfo {author} {\bibfnamefont {Steven~R.}\
  \bibnamefont {White}},\ }\bibfield  {title} {\enquote {\bibinfo {title}
  {Minimally entangled typical thermal state algorithms},}\ }\href@noop {}
  {\bibfield  {journal} {\bibinfo  {journal} {New Journal of Physics}\ }\textbf
  {\bibinfo {volume} {12}},\ \bibinfo {pages} {32} (\bibinfo {year}
  {2010})}\BibitemShut {NoStop}%
\bibitem [{\citenamefont {Evenbly}\ and\ \citenamefont
  {Vidal}(2011{\natexlab{b}})}]{Evenbly:2011}%
  \BibitemOpen
  \bibfield  {author} {\bibinfo {author} {\bibfnamefont {Glen}\ \bibnamefont
  {Evenbly}}\ and\ \bibinfo {author} {\bibfnamefont {Guifre}\ \bibnamefont
  {Vidal}},\ }\bibfield  {title} {\enquote {\bibinfo {title} {Quantum
  criticality with the multi-scale entanglement renormalization ansatz},}\
  }\href {http://arxiv.org/abs/1109.5334} {\bibfield  {journal} {\bibinfo
  {journal} {arxiv:1109.5334}\ } (\bibinfo {year} {2011}{\natexlab{b}})},\
  \Eprint {http://arxiv.org/abs/arxiv:1109.5334} {arxiv:1109.5334} \BibitemShut
  {NoStop}%
\bibitem [{ITe()}]{ITensor}%
  \BibitemOpen
  \href@noop {} {\bibinfo  {journal} {\mbox{ITensor Library} (version 2.0.7)
  http://itensor.org}\ }\BibitemShut {NoStop}%
\bibitem [{\citenamefont {Sch{\"o}lkopf}\ \emph {et~al.}(2001)\citenamefont
  {Sch{\"o}lkopf}, \citenamefont {Herbrich},\ and\ \citenamefont
  {Smola}}]{Scholkopf:2001}%
  \BibitemOpen
\bibfield  {journal} {  }\bibfield  {author} {\bibinfo {author} {\bibfnamefont
  {Bernhard}\ \bibnamefont {Sch{\"o}lkopf}}, \bibinfo {author} {\bibfnamefont
  {Ralf}\ \bibnamefont {Herbrich}}, \ and\ \bibinfo {author} {\bibfnamefont
  {Alex}\ \bibnamefont {Smola}},\ }\bibfield  {title} {\enquote {\bibinfo
  {title} {A generalized representer theorem},}\ }in\ \href@noop {} {\emph
  {\bibinfo {booktitle} {Computational learning theory}}}\ (\bibinfo
  {organization} {Springer},\ \bibinfo {year} {2001})\ pp.\ \bibinfo {pages}
  {416--426}\BibitemShut {NoStop}%
\bibitem [{\citenamefont {Sch{\"o}lkopf}\ \emph {et~al.}(1998)\citenamefont
  {Sch{\"o}lkopf}, \citenamefont {Smola},\ and\ \citenamefont
  {M{\"u}ller}}]{Scholkopf:1998}%
  \BibitemOpen
  \bibfield  {author} {\bibinfo {author} {\bibfnamefont {Bernhard}\
  \bibnamefont {Sch{\"o}lkopf}}, \bibinfo {author} {\bibfnamefont {Alexander}\
  \bibnamefont {Smola}}, \ and\ \bibinfo {author} {\bibfnamefont
  {Klaus-Robert}\ \bibnamefont {M{\"u}ller}},\ }\bibfield  {title} {\enquote
  {\bibinfo {title} {Nonlinear component analysis as a kernel eigenvalue
  problem},}\ }\href@noop {} {\bibfield  {journal} {\bibinfo  {journal} {Neural
  Computation}\ }\textbf {\bibinfo {volume} {10}},\ \bibinfo {pages}
  {1299--1319} (\bibinfo {year} {1998})}\BibitemShut {NoStop}%
\bibitem [{\citenamefont {Ba{\~n}uls}\ \emph {et~al.}(2008)\citenamefont
  {Ba{\~n}uls}, \citenamefont {P{\'e}rez-Garc{\'\i}a}, \citenamefont {Wolf},
  \citenamefont {Verstraete},\ and\ \citenamefont {Cirac}}]{Banuls:2008}%
  \BibitemOpen
  \bibfield  {author} {\bibinfo {author} {\bibfnamefont {M.}~\bibnamefont
  {Ba{\~n}uls}}, \bibinfo {author} {\bibfnamefont {D.}~\bibnamefont
  {P{\'e}rez-Garc{\'\i}a}}, \bibinfo {author} {\bibfnamefont {M.M.}\
  \bibnamefont {Wolf}}, \bibinfo {author} {\bibfnamefont {F.}~\bibnamefont
  {Verstraete}}, \ and\ \bibinfo {author} {\bibfnamefont {J.I.}\ \bibnamefont
  {Cirac}},\ }\bibfield  {title} {\enquote {\bibinfo {title} {Sequentially
  generated states for the study of two-dimensional systems},}\ }\href@noop {}
  {\bibfield  {journal} {\bibinfo  {journal} {Phys. Rev. A}\ }\textbf {\bibinfo
  {volume} {77}} (\bibinfo {year} {2008})}\BibitemShut {NoStop}%
\bibitem [{\citenamefont {Oseledets}(2011)}]{Oseledets:2011}%
  \BibitemOpen
  \bibfield  {author} {\bibinfo {author} {\bibfnamefont {I.}~\bibnamefont
  {Oseledets}},\ }\bibfield  {title} {\enquote {\bibinfo {title} {Tensor-train
  decomposition},}\ }\href@noop {} {\bibfield  {journal} {\bibinfo  {journal}
  {SIAM Journal on Scientific Computing}\ }\textbf {\bibinfo {volume} {33}},\
  \bibinfo {pages} {2295--2317} (\bibinfo {year} {2011})}\BibitemShut {NoStop}%
\bibitem [{\citenamefont {Friedman}(1997)}]{Friedman:1997}%
  \BibitemOpen
  \bibfield  {author} {\bibinfo {author} {\bibfnamefont {Barry}\ \bibnamefont
  {Friedman}},\ }\bibfield  {title} {\enquote {\bibinfo {title} {A density
  matrix renormalization group approach to interacting quantum systems on
  cayley trees},}\ }\href@noop {} {\bibfield  {journal} {\bibinfo  {journal}
  {Journal of Physics: Condensed Matter}\ }\textbf {\bibinfo {volume} {9}},\
  \bibinfo {pages} {9021} (\bibinfo {year} {1997})}\BibitemShut {NoStop}%
\bibitem [{\citenamefont {Shi}\ \emph {et~al.}(2006)\citenamefont {Shi},
  \citenamefont {Duan},\ and\ \citenamefont {Vidal}}]{Shi:2006}%
  \BibitemOpen
  \bibfield  {author} {\bibinfo {author} {\bibfnamefont {Y.-Y.}\ \bibnamefont
  {Shi}}, \bibinfo {author} {\bibfnamefont {L.-M.}\ \bibnamefont {Duan}}, \
  and\ \bibinfo {author} {\bibfnamefont {G.}~\bibnamefont {Vidal}},\ }\bibfield
   {title} {\enquote {\bibinfo {title} {Classical simulation of quantum
  many-body systems with a tree tensor network},}\ }\href@noop {} {\bibfield
  {journal} {\bibinfo  {journal} {Phys. Rev. A}\ }\textbf {\bibinfo {volume}
  {74}},\ \bibinfo {pages} {022320} (\bibinfo {year} {2006})}\BibitemShut
  {NoStop}%
\bibitem [{\citenamefont {Hackbusch}\ and\ \citenamefont
  {K{\"u}hn}(2009)}]{Hackbusch:2009}%
  \BibitemOpen
  \bibfield  {author} {\bibinfo {author} {\bibfnamefont {W.}~\bibnamefont
  {Hackbusch}}\ and\ \bibinfo {author} {\bibfnamefont {S.}~\bibnamefont
  {K{\"u}hn}},\ }\bibfield  {title} {\enquote {\bibinfo {title} {A new scheme
  for the tensor representation},}\ }\href@noop {} {\bibfield  {journal}
  {\bibinfo  {journal} {Journal of Fourier Analysis and Applications}\ }\textbf
  {\bibinfo {volume} {15}},\ \bibinfo {pages} {706--722} (\bibinfo {year}
  {2009})}\BibitemShut {NoStop}%
\bibitem [{\citenamefont {Cichocki}(2014)}]{Cichocki:2014}%
  \BibitemOpen
  \bibfield  {author} {\bibinfo {author} {\bibfnamefont {Andrzej}\ \bibnamefont
  {Cichocki}},\ }\bibfield  {title} {\enquote {\bibinfo {title} {Era of big
  data processing: A new approach via tensor networks and tensor
  decompositions},}\ }\href@noop {} {\bibfield  {journal} {\bibinfo  {journal}
  {arxiv:1403.2048}\ } (\bibinfo {year} {2014})}\BibitemShut {NoStop}%
\bibitem [{\citenamefont {Bridgeman}\ and\ \citenamefont
  {Chubb}(2016)}]{Bridgeman:2016}%
  \BibitemOpen
  \bibfield  {author} {\bibinfo {author} {\bibfnamefont {Jacob~C.}\
  \bibnamefont {Bridgeman}}\ and\ \bibinfo {author} {\bibfnamefont
  {Christopher~T.}\ \bibnamefont {Chubb}},\ }\bibfield  {title} {\enquote
  {\bibinfo {title} {Hand-waving and interpretive dance: An introductory course
  on tensor networks},}\ }\href@noop {} {\bibfield  {journal} {\bibinfo
  {journal} {arxiv:1603.03039}\ } (\bibinfo {year} {2016})}\BibitemShut
  {NoStop}%
\bibitem [{\citenamefont {Silvi}\ \emph {et~al.}(2017)\citenamefont {Silvi},
  \citenamefont {Tschirsich}, \citenamefont {Gerster}, \citenamefont
  {J{\"u}nemann}, \citenamefont {Jaschke}, \citenamefont {Rizzi},\ and\
  \citenamefont {Montangero}}]{Silvi:2017}%
  \BibitemOpen
  \bibfield  {author} {\bibinfo {author} {\bibfnamefont {Pietro}\ \bibnamefont
  {Silvi}}, \bibinfo {author} {\bibfnamefont {Ferdinand}\ \bibnamefont
  {Tschirsich}}, \bibinfo {author} {\bibfnamefont {Matthias}\ \bibnamefont
  {Gerster}}, \bibinfo {author} {\bibfnamefont {Johannes}\ \bibnamefont
  {J{\"u}nemann}}, \bibinfo {author} {\bibfnamefont {Daniel}\ \bibnamefont
  {Jaschke}}, \bibinfo {author} {\bibfnamefont {Matteo}\ \bibnamefont {Rizzi}},
  \ and\ \bibinfo {author} {\bibfnamefont {Simone}\ \bibnamefont
  {Montangero}},\ }\bibfield  {title} {\enquote {\bibinfo {title} {The tensor
  networks anthology: Simulation techniques for many-body quantum lattice
  systems},}\ }\href@noop {} {\bibfield  {journal} {\bibinfo  {journal}
  {arxiv:1710.03733}\ } (\bibinfo {year} {2017})}\BibitemShut {NoStop}%
\bibitem [{\citenamefont {Yann~LeCun}()}]{MNIST}%
  \BibitemOpen
  \bibfield  {author} {\bibinfo {author} {\bibfnamefont {Christopher
  J.C.~Burges}\ \bibnamefont {Yann~LeCun}, \bibfnamefont {Corinna~Cortes}},\
  }\bibfield  {title} {\enquote {\bibinfo {title} {\mbox{MNIST} handwritten
  digit database},}\ }\href {http://yann.lecun.com/exdb/mnist/} {\bibinfo
  {journal} {http://yann.lecun.com/exdb/mnist/}\ }\BibitemShut {NoStop}%
\bibitem [{\citenamefont {Xiao}\ \emph {et~al.}(2017)\citenamefont {Xiao},
  \citenamefont {Rasul},\ and\ \citenamefont {Vollgraf}}]{Xiao:2017}%
  \BibitemOpen
\bibfield  {journal} {  }\bibfield  {author} {\bibinfo {author} {\bibfnamefont
  {Han}\ \bibnamefont {Xiao}}, \bibinfo {author} {\bibfnamefont {Kashif}\
  \bibnamefont {Rasul}}, \ and\ \bibinfo {author} {\bibfnamefont {Roland}\
  \bibnamefont {Vollgraf}},\ }\bibfield  {title} {\enquote {\bibinfo {title}
  {Fashion-\mbox{MNIST}: a novel image dataset for benchmarking machine
  learning algorithms},}\ }\href@noop {} {\bibfield  {journal} {\bibinfo
  {journal} {arxiv:1708.07747}\ } (\bibinfo {year} {2017})}\BibitemShut
  {NoStop}%
\bibitem [{\citenamefont {Scholkopf}\ and\ \citenamefont
  {Smola}(2001)}]{Scholkopf:2001L}%
  \BibitemOpen
  \bibfield  {author} {\bibinfo {author} {\bibfnamefont {B.}~\bibnamefont
  {Scholkopf}}\ and\ \bibinfo {author} {\bibfnamefont {A.}~\bibnamefont
  {Smola}},\ }\href@noop {} {\emph {\bibinfo {title} {Learning with kernels:
  Support Vector Machines, regularization, optimization, and beyond}}}\
  (\bibinfo  {publisher} {MIT Press},\ \bibinfo {year} {2001})\BibitemShut
  {NoStop}%
\bibitem [{\citenamefont {Bach}(2013)}]{Bach:2013}%
  \BibitemOpen
  \bibfield  {author} {\bibinfo {author} {\bibfnamefont {F.}~\bibnamefont
  {Bach}},\ }\bibfield  {title} {\enquote {\bibinfo {title} {Sharp analysis of
  low-rank kernel matrix approximations},}\ }in\ \href@noop {} {\emph {\bibinfo
  {booktitle} {Proceedings of COLT}}}\ (\bibinfo {year} {2013})\BibitemShut
  {NoStop}%
\bibitem [{\citenamefont {Cesa-Bianchi}\ \emph {et~al.}(2015)\citenamefont
  {Cesa-Bianchi}, \citenamefont {Mansour},\ and\ \citenamefont
  {Shamir}}]{Cesa-Bianchi}%
  \BibitemOpen
  \bibfield  {author} {\bibinfo {author} {\bibfnamefont {N.}~\bibnamefont
  {Cesa-Bianchi}}, \bibinfo {author} {\bibfnamefont {Y.}~\bibnamefont
  {Mansour}}, \ and\ \bibinfo {author} {\bibfnamefont {O.}~\bibnamefont
  {Shamir}},\ }\bibfield  {title} {\enquote {\bibinfo {title} {On the
  complexity of learning with kernels},}\ }\href@noop {} {\bibfield  {journal}
  {\bibinfo  {journal} {Proceedings of The 28th Conference on Learning Theory}\
  ,\ \bibinfo {pages} {297--325}} (\bibinfo {year} {2015})}\BibitemShut
  {NoStop}%
\bibitem [{\citenamefont {Rahimi}\ and\ \citenamefont
  {Recht}(2008)}]{Rahimi:2008}%
  \BibitemOpen
  \bibfield  {author} {\bibinfo {author} {\bibfnamefont {Ali}\ \bibnamefont
  {Rahimi}}\ and\ \bibinfo {author} {\bibfnamefont {Benjamin}\ \bibnamefont
  {Recht}},\ }\bibfield  {title} {\enquote {\bibinfo {title} {Random features
  for large-scale kernel machines},}\ }in\ \href@noop {} {\emph {\bibinfo
  {booktitle} {Advances in Neural Information Processing Systems 20}}},\
  \bibinfo {editor} {edited by\ \bibinfo {editor} {\bibfnamefont {J.~C.}\
  \bibnamefont {Platt}}, \bibinfo {editor} {\bibfnamefont {D.}~\bibnamefont
  {Koller}}, \bibinfo {editor} {\bibfnamefont {Y.}~\bibnamefont {Singer}}, \
  and\ \bibinfo {editor} {\bibfnamefont {S.~T.}\ \bibnamefont {Roweis}}}\
  (\bibinfo  {publisher} {Curran Associates, Inc.},\ \bibinfo {year} {2008})\
  pp.\ \bibinfo {pages} {1177--1184}\BibitemShut {NoStop}%
\bibitem [{\citenamefont {Rudi}\ \emph {et~al.}(2017)\citenamefont {Rudi},
  \citenamefont {Carratino},\ and\ \citenamefont {Rosasco}}]{Rudi:2017}%
  \BibitemOpen
  \bibfield  {author} {\bibinfo {author} {\bibfnamefont {Alessandro}\
  \bibnamefont {Rudi}}, \bibinfo {author} {\bibfnamefont {Luigi}\ \bibnamefont
  {Carratino}}, \ and\ \bibinfo {author} {\bibfnamefont {Lorenzo}\ \bibnamefont
  {Rosasco}},\ }\bibfield  {title} {\enquote {\bibinfo {title} {\mbox{FALKON}:
  An optimal large scale kernel method},}\ }\href@noop {} {\bibfield  {journal}
  {\bibinfo  {journal} {arxiv:1705.10958}\ } (\bibinfo {year}
  {2017})}\BibitemShut {NoStop}%
\bibitem [{\citenamefont {Vidal}(2007{\natexlab{b}})}]{Vidal:2007a}%
  \BibitemOpen
  \bibfield  {author} {\bibinfo {author} {\bibfnamefont {G.}~\bibnamefont
  {Vidal}},\ }\bibfield  {title} {\enquote {\bibinfo {title} {Algorithms for
  entanglement renormalization},}\ }\href@noop {} {\bibfield  {journal}
  {\bibinfo  {journal} {https://arxiv.org/abs/0707.1454v2}\ } (\bibinfo {year}
  {2007}{\natexlab{b}})}\BibitemShut {NoStop}%
\bibitem [{\citenamefont {Dolfi}\ \emph {et~al.}(2012)\citenamefont {Dolfi},
  \citenamefont {Bauer}, \citenamefont {Troyer},\ and\ \citenamefont
  {Ristivojevic}}]{Dolfi:2012}%
  \BibitemOpen
  \bibfield  {author} {\bibinfo {author} {\bibfnamefont {Michele}\ \bibnamefont
  {Dolfi}}, \bibinfo {author} {\bibfnamefont {Bela}\ \bibnamefont {Bauer}},
  \bibinfo {author} {\bibfnamefont {Matthias}\ \bibnamefont {Troyer}}, \ and\
  \bibinfo {author} {\bibfnamefont {Zoran}\ \bibnamefont {Ristivojevic}},\
  }\bibfield  {title} {\enquote {\bibinfo {title} {Multigrid algorithms for
  tensor network states},}\ }\href@noop {} {\bibfield  {journal} {\bibinfo
  {journal} {Phys. Rev. Lett.}\ }\textbf {\bibinfo {volume} {109}},\ \bibinfo
  {pages} {020604} (\bibinfo {year} {2012})}\BibitemShut {NoStop}%
\bibitem [{\citenamefont {Cramer}\ \emph {et~al.}(2010)\citenamefont {Cramer},
  \citenamefont {Plenio}, \citenamefont {Flammia}, \citenamefont {Somma},
  \citenamefont {Gross}, \citenamefont {Bartlett}, \citenamefont
  {Landon-Cardinal}, \citenamefont {Poulin},\ and\ \citenamefont
  {Liu}}]{Cramer:2010}%
  \BibitemOpen
  \bibfield  {author} {\bibinfo {author} {\bibfnamefont {Marcus}\ \bibnamefont
  {Cramer}}, \bibinfo {author} {\bibfnamefont {Martin~B.}\ \bibnamefont
  {Plenio}}, \bibinfo {author} {\bibfnamefont {Steven~T.}\ \bibnamefont
  {Flammia}}, \bibinfo {author} {\bibfnamefont {Rolando}\ \bibnamefont
  {Somma}}, \bibinfo {author} {\bibfnamefont {David}\ \bibnamefont {Gross}},
  \bibinfo {author} {\bibfnamefont {Stephen~D.}\ \bibnamefont {Bartlett}},
  \bibinfo {author} {\bibfnamefont {Olivier}\ \bibnamefont {Landon-Cardinal}},
  \bibinfo {author} {\bibfnamefont {David}\ \bibnamefont {Poulin}}, \ and\
  \bibinfo {author} {\bibfnamefont {Yi-Kai}\ \bibnamefont {Liu}},\ }\bibfield
  {title} {\enquote {\bibinfo {title} {Efficient quantum state tomography},}\
  }\href@noop {} {\bibfield  {journal} {\bibinfo  {journal} {Nature
  Communications}\ }\textbf {\bibinfo {volume} {1}},\ \bibinfo {pages} {149 EP
  --} (\bibinfo {year} {2010})}\BibitemShut {NoStop}%
\bibitem [{\citenamefont {Landon-Cardinal}\ and\ \citenamefont
  {Poulin}(2012)}]{Landon-Cardinal:2012}%
  \BibitemOpen
  \bibfield  {author} {\bibinfo {author} {\bibfnamefont {Olivier}\ \bibnamefont
  {Landon-Cardinal}}\ and\ \bibinfo {author} {\bibfnamefont {David}\
  \bibnamefont {Poulin}},\ }\bibfield  {title} {\enquote {\bibinfo {title}
  {Practical learning method for multi-scale entangled states},}\ }\href@noop
  {} {\bibfield  {journal} {\bibinfo  {journal} {New Journal of Physics}\
  }\textbf {\bibinfo {volume} {14}},\ \bibinfo {pages} {085004} (\bibinfo
  {year} {2012})}\BibitemShut {NoStop}%
\bibitem [{\citenamefont {Nouy}(2017)}]{Nouy:2017}%
  \BibitemOpen
  \bibfield  {author} {\bibinfo {author} {\bibfnamefont {Anthony}\ \bibnamefont
  {Nouy}},\ }\bibfield  {title} {\enquote {\bibinfo {title} {Higher-order
  principal component analysis for the approximation of tensors in tree-based
  low-rank formats},}\ }\href@noop {} {\bibfield  {journal} {\bibinfo
  {journal} {arxiv:1705.00880}\ } (\bibinfo {year} {2017})}\BibitemShut
  {NoStop}%
\end{thebibliography}%

\end{document}